\documentclass[a4paper]{paper_cw}

\usepackage{packages}

\renewbibmacro*{title}{%
  \ifentrytype{misc}
    {\printtext{\textit{\thefield{title}}}} 
    {\ifentrytype{book}
      {\printtext{\textit{\thefield{title}}}} 
      {\printtext[title]{\thefield{title}}} 
    }
}

\renewbibmacro*{booktitle}{%
\printtext{\textit{\thefield{booktitle}}} 
}
\DeclareFieldFormat[article]{journal}{\textit{#1}}

\begin{document}

\title{Improved Training Strategies for Physics-Informed Neural Networks using Real Experimental Data in Aluminum Spot Welding}

\author{Jan A. Zak$^1$, Christian Weißenfels$^{1,2}$}
 
\institute{Jan A. Zak (\Letter)
\\\email{jan1.zak@uni-a.de}\\
$^1$University of Augsburg, Germany\\
$^2$Center of Advanced Analytics and Predictive Sciences, University of Augsburg, Germany
}

\maketitle
\thispagestyle{empty}

\abstract{
Resistance spot welding is the dominant joining process for the body-in-white in the automotive industry, where the weld nugget diameter is the key quality metric. Its measurement requires destructive testing, limiting the potential for efficient quality control. 
Physics-informed neural networks were investigated as a promising tool to reconstruct internal process states from experimental data, enabling model-based and non-invasive quality assessment in aluminum spot welding.
A major challenge is the integration of real-world data into the network due to competing optimization objectives. To address this, we introduce two novel training strategies. First, experimental losses for dynamic displacement and nugget diameter are progressively included using a fading-in function to prevent excessive optimization conflicts. We also implement a custom learning rate scheduler and early stopping based on a rolling window to counteract premature reduction due to increased loss magnitudes.
Second, we introduce a conditional update of temperature-dependent material parameters via a look-up table, activated only after a loss threshold is reached to ensure physically meaningful temperatures.
An axially symmetric two-dimensional model was selected to represent the welding process accurately while maintaining computational efficiency. To reduce computational burden, the training strategies and model components were first systematically evaluated in one dimension, enabling controlled analysis of loss design and contact models. The two-dimensional network predicts dynamic displacement and nugget growth within the experimental confidence interval, supports transferring welding stages from steel to aluminum, and demonstrates strong potential for fast, model-based quality control in industrial applications.
}

\keywords{physics-informed neural networks \and data-driven modeling \and aluminum resistance spot welding \and dynamic displacement \and weld nugget diameter}

\section{Introduction}
In recent years, automotive manufacturers have prioritized reducing vehicle weight \autocite{Eckstein2011,Barnes2000a} due to regulatory scrutiny on $\CO$ emissions from vehicles \autocite{EU2019} or increasing safety requirements. Around a quarter of the vehicle's total weight is constituted by the \biw\ \autocite{Roth2001}; the stage of a car body's frame before it has been painted and the installation of components like electrical wiring, powertrain, and trim.
New construction methods, such as lightweight construction, are used to reduce weight. One competitive substitution for relatively weighty steel are \acrlong{Al} alloys due to their high \mbox{strength-to-weight} ratio. This transition results in different materials with distinct properties, requiring the development of proper joining technologies. For example, \acrlong{Al} exhibits approximately \mbox{one-third} the electric resistivity and three times the thermal conductivity of \acrlong{Fe} at room temperature \autocite{Gould2012}. Furthermore, the liquidus temperature is 660 degrees \celsius, in contrast to 1,500 degrees for steel.

%
%
The dominant joining technology for the \biw\ is \acrfull{RSW}. It is an established process to join metal sheets in the automotive industry and other sectors, such as aerospace \autocite{Dobler1969}. \acrshort{RSW} joins two surfaces with heat generated by the resistance to the flow of electric current \autocite{ASM2011}. The contact between the surfaces is created by local clamping of the workpieces with two electrodes, as illustrated in \fig~\ref{fig:RSW}.
%
%
RSW of steel has been the focus of various studies investigating influencing factors, such as electrode force \autocite{Ao2009,Furukawa2006b}, surface conditions \autocite{Gould1987,Bowden1958}, electrical contact behavior \autocite{Wei2012,Wei2014}, and thermal phenomena \autocite{Hulst1969,Greenwood1958}. These investigations have primarily focused on either experimental approaches, which emphasize process parameters and material properties, or on the development of numerical models, mainly using the \acrfull{FEM}. Experimental investigations focused on various aspects, such as electrode cap lifetime \autocite{Boomer2003}, electrode cap form \autocite{Bowers1990}, or specific material combinations. \textcite{Wei1996} improved the modeling of nugget formation for steel \acrshort{RSW} by integrating a current density model with experimental data from \textcite{Gould1987}. Another study by \textcite{Wang2015} validated a numerical model for \mbox{\acrlong{Al}-steel} combinations using experimental measurements. Similarly, \textcite{Prabitz2021} developed and parameterized a numerical model specifically for ultra-high-strength steel. These studies underscore the critical role of both experimental and numerical approaches in enhancing the understanding of \acrshort{RSW}, particularly regarding different materials and optimizing welding parameters.
The primary quality measure is the nugget diameter $\nuggetDia \ [\nuggetDiaUnit]$, proportional to the joint strength \autocite{ISO14270}.
The nugget diameter of steel \acrshort{RSW} has been modeled using two-dimensional axisymmetric \acrshort{FEM} simulations.
\textcite{Hou2007} used ANSYS, with results including temperature profiles, the heat-affected zone, and electrode displacement. The authors validated the results against the simulation results from \textcite{Tsai1991}, showing a good approximation.
\textcite{Eisazadeh2010} developed a model using ANSYS to predict nugget thickness development directly, validating their results against experimental data from \textcite{Gould1987}.
\textcite{Saleem2012} used COMSOL to investigate the influence of electrode tip area on the welding process across three distinct cases, providing insights into how geometric parameters influence nugget formation \textcite{Hou2007}.
Experimental studies have been conducted by \textcite{Gedeon1986b} and \textcite{Lane1987}. \citeauthor{Gedeon1986b} data was collected on galvanized steel, which is welded with lower currents, a 133 ms upslope, and 266 ms of main weld. The authors conducted a series of experiments in which the welding process was halted at predetermined intervals to measure $\nuggetDia$.

%
%
In contrast to the extensive research on steel \acrshort{RSW}, studies on \acrlong{Al} alloys are comparatively limited, particularly in the context of \mbox{large-scale} production and big data. 
\textcite{Florea2013} conducted an experimental study using a \mbox{trial-and-error} approach to identify optimal process parameters, primarily focusing on maximizing electrical current for welding \mbox{2.0-\mm} \AlSZSO. \textcite{Ji2010} investigated the dynamic displacement across varying process parameters and postulated a linear relationship between the maximum displacement and nugget diameter. \textcite{Hu2022} utilized metrics, including the peak value of dynamic displacement, to predict weld quality. The authors used ensemble models and concluded that displacement is a good predictor.
\textcite{Kim2018} conducted experiments and simulations using commercial welding software to investigate the effect of welding time on a \mbox{1.2-\mm} sheet of \AlFZFT. The simulation and experimental measurements indicate that the contact surface surpasses the liquidus temperature within the first 33.4 \ms\ of the main weld time.

%
%
Traditional physics-based approaches, such as the \acrshort{FEM}, can solve complex physical problems accurately when parameters are well-defined. However, even with experimental measurements used for calibration and validation, FEM possesses limitations in its capacity to learn and incorporate experimental data.
These data integration challenges can be addressed with Physics-informed neural networks (PINNs) \autocite{Lagaris1998,Karniadakis2021}. They integrate partial differential equations (PDEs) into a \acrlong{NN} by regularizing for physical consistency. PINNs provide a mesh-free solution \autocite{Cuomo2022} and can alleviate the curse of dimensionality \autocite{Poggio2017,Grohs2023}. Moreover, they can solve inverse problems, allowing for the direct inference of unknown parameters based on experimental data.
PINNs enable the direct incorporation of experimental measurements into the computation of the solution function for initial boundary value problems. In contrast, FEM solvers typically require additional inverse procedures to infer unknown parameters from data. However, the use of PINNs typically requires significant computational resources during training \autocite{Grossmann2023} and presents challenges in hyperparameter selection.
Various scientific and engineering problems that involve \acrshort{PDE}s have been studied, \eg\ fluid mechanics \autocite{Wessels2020,Sun2020} or materials science \autocite{Zhang2022PINN,Chen2020}.

%
%
\textcite{Kapoor2024} utilized \acrshort{PINN}s to solve known analytical solutions for beam bending moments, successfully identifying unknown forces. \textcite{Moseley2020} solved the wave equation by augmenting the loss function with terms derived from the respective \acrshort{PDE}.
\textcite{Hennigh2021} derived the geometry of a heat sink purely based on the governing \acrshort{PDE}.
Similarly, \textcite{Pulido2022} developed a \acrlong{2D} \acrshort{PINN} to address a magnetostatic problem, validating the results with a \acrshort{FEM} solver.
\textcite{Wang2023Scale} developed a \acrshort{PINN} for \acrlong{2D} ultrasound wave propagation, tackling the challenges of numerical solvers with discrete boundary conditions. 
\textcite{Cai2022} provided an extensive review of \mbox{three-dimensional} fluid dynamics applications using \acrshort{PINN}s and further explored the topic by conducting experiments on heat flow over an espresso cup. Their work highlighted the benefits of integrating experimental data, as they inferred velocity and pressure fields from temperature data, demonstrating how experimental inputs can enhance the training and accuracy of \acrshort{PINN}s.

%
%
\acrshort{PINN}s in industrial contexts remain limited, particularly in \acrshort{RSW}. We adopt a data-driven approach and validate the predictions against experimental measurements. To this end, we divide the data into training, validation, and test sets. We develop two setups for a \acrshort{1D} and \acrshort{2D} model that integrate experimental data with governing equations to solve the temperature and displacement fields. We develop training strategies for fast loss convergence, resolving optimization challenges using experimental data, and develop a \temperatureDependent\ material update procedure. This is a challenging task, which is achieved by delaying the activation of experimental loss and developing a rolling window for the learning rate scheduler, as well as implementing early stopping. Furthermore, we incorporate the RSW-specific welding schedule, contact heat generation, and the welding control behavior to capture the dynamic displacement curve. The resulting temperature field directly predicts the dynamic displacement and the nugget diameter, the primary quality measure.

\section{Resistance Spot Welding}
\begin{figure}
    \centering
    \includegraphics[width=\textwidth]{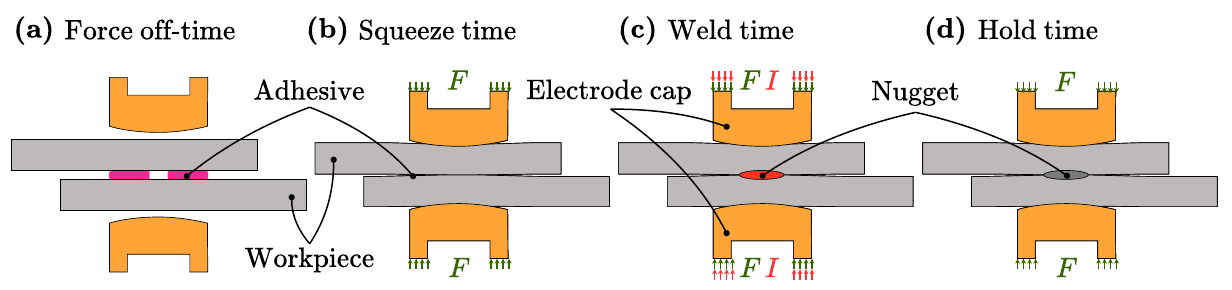}
    \vspace*{-2mm}
    \caption{The \acrshort{RSW} phases encompass the initial setup, squeezing the workpieces together, executing the weld, and holding the electrodes in place until they return to the initial setup.}
    \label{fig:RSW}
\end{figure}
In industrial applications, \acrshort{RSW} is often combined with adhesive bonding \autocite{Adams2021}. This process is generally divided into four phases \autocite{ISO17677}. The initial phase involves preparing the sheets and the application of the adhesive. During the second phase, known as squeeze time $\zeitS$, electrode caps contact the sheets and exert an electrode force $\forceMax \ [\forceMaxUnit]$. In the third phase, an electric current $\currentMax \ [\currentMaxUnit]$ is applied through the electrodes following a welding schedule. This causes the material to heat up due to \joule\ heating, forming a weld nugget between the sheets. The final phase, hold-time $\zeitH$, maintains the electrode force. The process is illustrated in \fig~\ref{fig:RSW}.
Heat generation primarily occurs at the faying surface between the sheets and is amplified by the adhesive. Heat at the \electrodeSheet\ interface is managed through electrode cap cooling with $\coolTemp \ [\temperatureUnit]$. For steel, it is estimated that most of the generated heat is dissipated through electrode cooling and conduction into the sheets \autocite{Eagar1990}, and approximately 20\perc\ contributes to the nugget formation.

\subsection{Experimental setup}
Two \acrlong{AL6} with a thickness of 1.5 mm were used in our experiments. The chemical composition is listed in \tab~\ref{tab:chemische_zusammensetzung}.
The experimental design factors are $\currentMax$ and $\forceMax$, and the response variable is $\nuggetDia$. Four forces and 22 current factor levels result in 88 factor combinations. Each combination has been welded 30 times, yielding an experimental mean and standard deviation.
The welding schedule consists of a preheat current during $\zeitP$ of 12 kiloamperes over 400 \ms, a linear upslope of 70 \ms, and a main welding time of 90 \ms\ at $\currentMax$.
%
%
The welding schedule defines a current density $\currentDensity=\currentMax \apparentArea^{-1}$, where $\apparentArea$ is the apparent contact area between the sheet and electrode tip, during main welding and a constant heat source with current density $\currentDensity_\text{pt}$ during preheating. A generic schedule can be derived:
\begin{align}
    \currentDensity( \zeit ) &= \left\{\begin{array}{cl}
        \currentDensity_\text{pt} & : \ 0 \le \zeit \le 400 \\
        \currentDensity_\text{pt} + (\currentDensity - \currentDensity_\text{pt}) \times \frac{\zeit - 400}{470 - 400} & : \  400 \lt \zeit \le 470 \\
        \currentDensity & : \ 470 \lt \zeit \le 560 \\
        0 & : \ 560 \lt \zeit \le 760 \\
    \end{array} \right. \label{eq:stromprofil}
\end{align}
The experiments have been conducted with a \cShaped\ transformer spot welding gun for \acrlong{Al} alloys. The electric current is supplied and rectified with a \mbox{medium-frequency} welding transformer and controlled with a constant current control. The welding control is a \faRexroth\ \steuerungGroup, holding the current $\processCurrent$ and force profile $\processForce$ according to the welding schedule defined in \tab~\ref{tab:versuchsplan}. The welding schedule incorporates a preheat time $\zeitP$ for surface preconditioning \autocite{Luo2015} and an upslope to ensure a homogeneous temperature rise \autocite{Lane1987,Gedeon1986a}.
The dimensions of the upper sheet and lower sheets are $\sheetXConst \times \sheetYConst$ \mms. The welds are placed in two rows from the 1\textsuperscript{st} to the 15\textsuperscript{th} weld spot and from the 16\textsuperscript{th} to the 30\textsuperscript{th} weld spot. The sheets are pretreated with a \TiZrLong\ and a \hotmelt.
A specific weld bonding $\betamate$ adhesive is automatically applied beforehand. The \capAlloyLong\ electrode caps \autocite{ISO5182} are of the type \capTypeConst\ \autocite{ISO5821} and are cooled constantly at $\coolTemp = 19$. The cap geometry is defined in \mms\ with a cap thickness of 9, cap radius of $\capRadConst$, and cap diameter of $\capDiaConst$. After welding, the sheets are peeled \autocite{ISO10447}, and the average $\nuggetDia$ is measured.
The time series $\posTS$ was collected at a sampling rate of 1 kHz. The displacement measurement is initialized such that the position is set to zero at the start of the welding schedule, ensuring a consistent reference for subsequent displacement changes during the weld execution.
%
%
\begin{table}
\centering
    \begin{tabular}{llllllllllllllll}
    \toprule
    \textbf{Si} & \textbf{Fe} & \textbf{Cu} & \textbf{Mn} & \textbf{Mg} & \textbf{Cr} & \textbf{Zn} & \textbf{Ti} & \textbf{Al} \\
    \midrule
    0.53 & 0.20 & 0.06 & 0.09 & 0.56 & 0.01 & 0.01 & 0.03 & 98.43 \\
    \bottomrule
    \end{tabular}
\caption{Chemical composition in percentage by weight for \acrlong{AL6} specimen.}
\label{tab:chemische_zusammensetzung}
\end{table}
%
\begin{table}[b]
\caption{Factor levels and controllable parameters.}
\centering
\begin{tabular}{lccccc}
    \toprule
    \textbf{Factor} & \textbf{Symbol} & \multicolumn{2}{c}{\textbf{Min./Max.}} & \textbf{Unit} & \textbf{Value range} \\
    \textbf{/ Parameter}  & & - & + &  \\
    \midrule
    max. current & $\currentMax$  & 26 & 47 & $\currentTSUnit$ & $\left\{ \currentMax \in \mathbb{N} | 26 \le \currentMax \le 47 \right\}$ \\
    max. force   & $\forceMax$    & 5  & 8  & kN & $\left\{ \forceMax \in \mathbb{N} | 5 \le \forceMax \le 8 \right\}$ \\
    \hline
    preheat & $\currentVs$ & \multicolumn{2}{c}{$\currentVsConst$}  & $\currentTSUnit$ & \{12\} \\
    squeeze time     & $\zeitS$    & \multicolumn{2}{c}{$\zeitSConst$} & $\zeitUnit$ & $\left\{ \zeitS \in \mathbb{Z} | -200 < \zeitS \le 0   \right\}$ \\
    preheat time     & $\zeitP$    & \multicolumn{2}{c}{$\zeitPConst$} & $\zeitUnit$ & $\left\{ \zeitP \in \mathbb{N} | 0    < \zeitP \le 400 \right\}$ \\
    upslope time & $\zeitU$    & \multicolumn{2}{c}{$\zeitUConst$} & $\zeitUnit$ & $\left\{ \zeitU \in \mathbb{N} | 400  < \zeitU \le 470 \right\}$ \\
    weld time     & $\zeitW$    & \multicolumn{2}{c}{$\zeitWConst$} & $\zeitUnit$ & $\left\{ \zeitW \in \mathbb{N} | 470  < \zeitW \le 560 \right\}$ \\
    hold time     & $\zeitH$    & \multicolumn{2}{c}{$\zeitHConst$} & $\zeitUnit$ & $\left\{ \zeitH \in \mathbb{N} | 560  < \zeitH \le 760 \right\}$ \\
    \bottomrule
\end{tabular}
\label{tab:versuchsplan}
\end{table}
\subsection{Dynamic electrode displacement}
\begin{figure}
    \centering
    \includegraphics[width=0.7\textwidth]{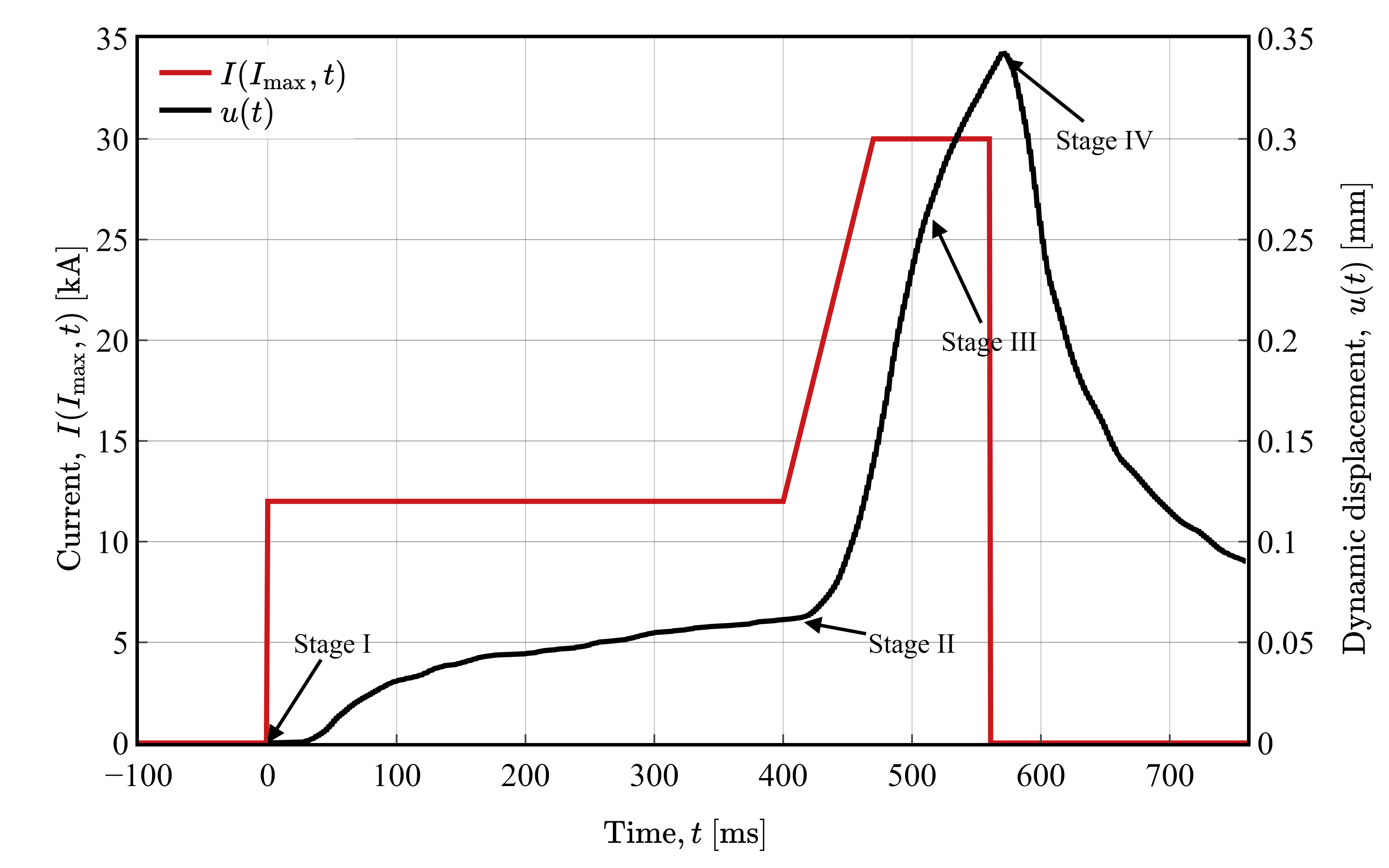}
    \caption{Welding schedule for current $I(\zeit)$ and dynamic displacement $\posTS(\zeit)$. The displacement is recorded up to 760 ms.}
    \label{fig:welding_schedule_disp}
\end{figure}
%
The process parameter $\currentMax$ results in a welding schedule $\processCurrent$. The force is typically held constant. An example welding schedule is illustrated in \fig~\ref{fig:welding_schedule_disp}. An example of the measured dynamic displacement is overlayed.
%
%
During welding, the sheet material thermally expands, which is measured as a dynamic displacement time series $\posTS(t) \ [\posTSUnit]$. The displacement is measured in the direction of the force application with an external sensor at the welding arm and quantifies the total expansion of the sheet material. The formation of the weld nugget can be monitored through dynamic displacement measurements during the welding process. 
\textcite{Gould1987} developed a one-dimensional finite difference thermal model for steel. 
Due to the \acrshort{1D} model, the author investigated the development of the nugget thickness, defined as the nugget height $\nuggetHeight$. The author mapped nugget height against welding current and time, identifying four growth stages: incubation, rapid growth, decreasing growth rate, and expulsion. While capturing the growth trends, the model overestimates nugget size in thicker sheets due to unaccounted radial heat flux.
\textcite{Gedeon1986b} investigated the development of nuggets and the dynamic monitoring of galvanized steel. The authors mapped stages of nugget development to $\posTS$. The study identifies six displacement curve regions, with two distinct stages for the zinc coating. Without those, the displacement curve can be classified into four regions: stage I initiates with an incubation period lasting approximately the first 33 \ms. No visible weld nugget growth is observed, while occasional local melting occurs. However, the surface asperities flatten, which may be observed by a decrease in $\posTS$.
Stage II progresses with the thermal expansion of the bulk material and rapid nugget growth. The thermal expansion rate is nearly linear until the metal begins to soften or reaches the melting point. \textcite{Dickinson1990} indicated a rapid growth originating from the periphery, forming an initial toroid shape that later transitions into an elliptical shape after about 50 \ms.
During stage III, the slope of $\posTS$ decreases once the metal softens. The weld nugget develops, and saturation occurs.
Stage IV is distinguished by a decrease or very slow growth of $\posTS$. Excessive heat generation during this phase or the previous may lead to spatter, an ejection of molten material.
\textcite{Wei1990} examined $\nuggetHeight$ growth in \acrshort{RSW} of steel using an axisymmetric heat conduction model. The authors employed an enthalpy method to account for phase changes due to melting and propose an effective model for heat generation at the faying surface, validated against experimental data from \textcite{Gould1987}.
\textcite{Ji2004} characterized the dynamic displacement and electrode force shape and investigated expulsion.

\subsection{Temperature-dependent material parameters}
\begin{figure}
    \centering
    \includegraphics[width=\textwidth]{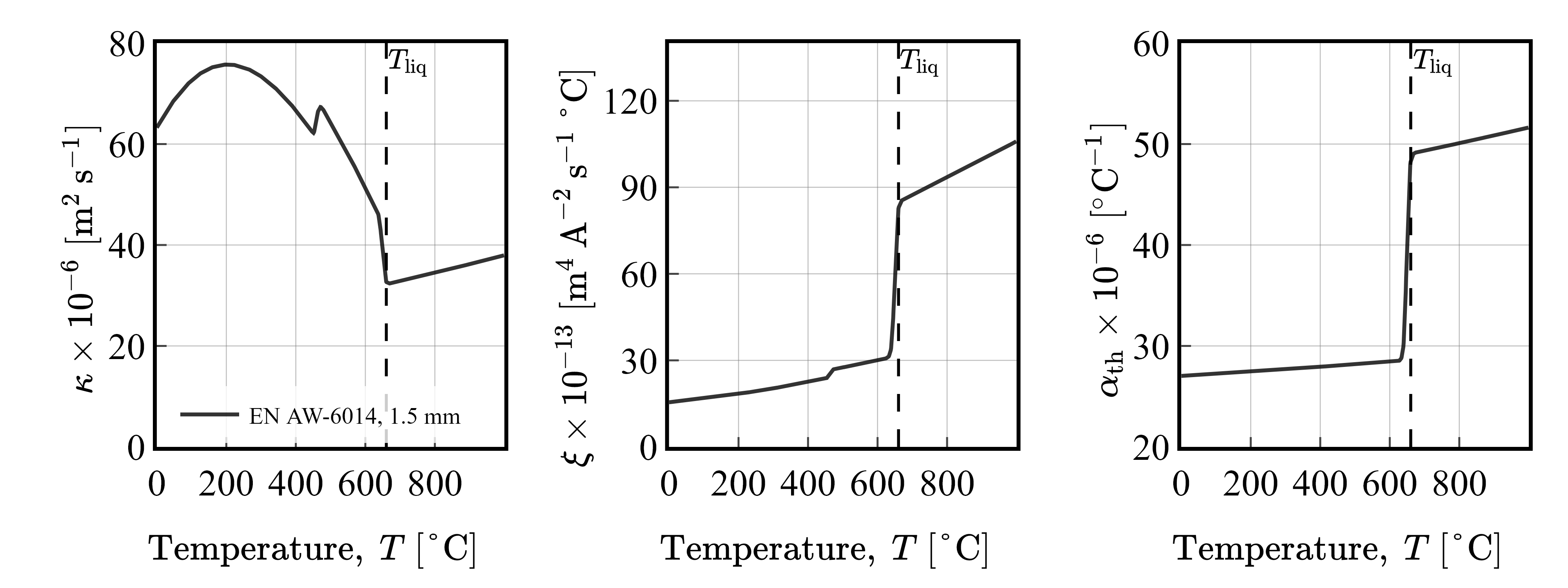}
    \caption{Experimentally measured \TemperatureDependent\ material parameters. Thermal diffusivity $\thermDiffusivity$, coefficient $\electrThermDissipation$, and thermal expansion $\thermExp$.}
    \label{fig:diffusivities}
\end{figure}
The material parameters of \acrlong{AL6} have been measured for the specific electrical resistance, density, and specific heat capacity up to 550 degrees \celsius. From those quantities, the thermal diffusivity $\thermDiffusivity \ [\thermDiffusivityUnit]$, the electric resistivity per heat capacity and density $\electrThermDissipation \ [\electrThermDissipationUnit]$, and the linear thermal expansion coefficient $\thermExp \ [\thermExpUnit]$ are defined. Values for higher temperatures and the liquidus were \mbox{curve-fitted} to the experimental measurements and sourced from \textcite{Touloukian1970a} for thermal conductivity, \textcite{Touloukian1970b} for specific heat capacity, \textcite{Mills2002} for density, \textcite{Touloukian1975} for thermal expansion, and \textcite{Leitner2017} for electrical resistance. The phase change from the solid to the liquid phase $\solTemp \rightarrow \liqTemp$ is modeled with the fraction of liquid material $\fracLiq (\temperature)$ \autocite{Kavallaris2018,Mills2002}. A piecewise function $f$ defines the phase transition, which is generically defined as:
\begin{align}
f (\temperature, \fracLiq) &= \left\{
\begin{array}{cl}
  f\phantom{}_\text{sol} & : \ \temperature \le \solTemp \\
  (1 - \fracLiq) f\phantom{}_\text{sol} + \fracLiq f\phantom{}_\text{liq} & : \ \solTemp \lt \temperature \lt \liqTemp \ \eqDot \\
  f\phantom{}_\text{liq} & : \ \temperature \ge \liqTemp \\
\end{array} \right. \label{eq:material_parameter}
\end{align}
The resulting thermophysical material parameters are depicted in \fig~\ref{fig:diffusivities}. Tensile tests have also been conducted up to 500 degrees, from which the hardness $\hardness(\temperature)$ is estimated to decrease linearly until $\liqTemp$.

\subsection{Welding control behavior}
\begin{figure}[]
    \centering
    \includegraphics[height=0.25\textheight]{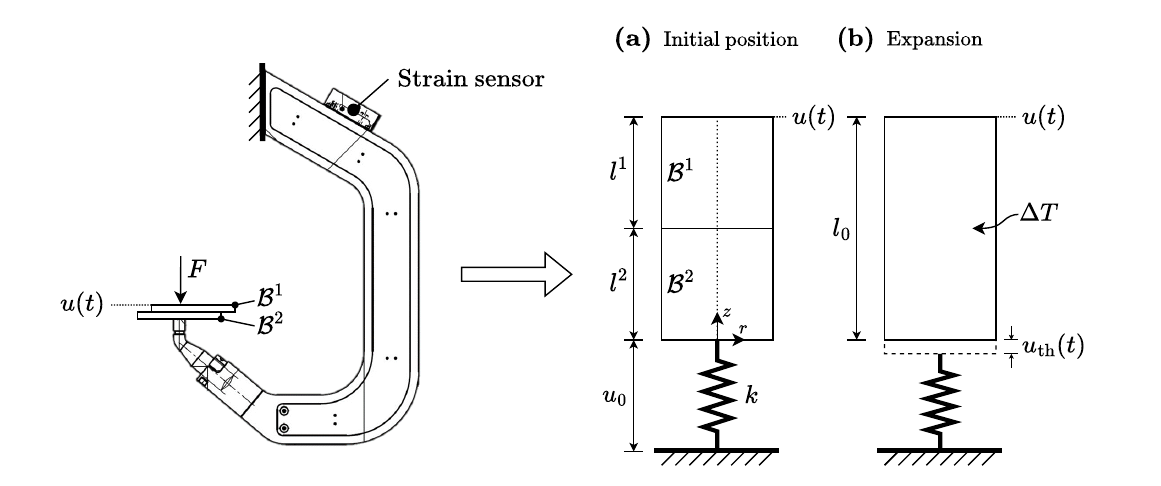}
    \caption{Simplified system model. Left: Welding gun with the application point of $\force$ and the position of the sheets. Right: Simplified \acrlong{2D} system with a spring displacement $\displacementSpring$ and the upper and lower sheet thicknesses $l^1$ and $l^2$, respectively.}
    \label{fig:ersatzsystem}
\end{figure}
The welding control uses an indirect measurement of the electrode force to minimize electromagnetic interference. Force is measured with a surface strain sensor on the static electrode \cShaped\ arm of the welding gun. It assumes linear elasticity and is calibrated against a force transducer.
The sensor leverages the gun structure to measure the displacement of the arm. The result of the control is a step response $\displacementSpring(t)$ to offset the current thermal expansion $\thermalDisplacement(t)$ of the sheets. Therefore, the measured dynamic displacement does not equal the thermal expansion, $\posTS(t) \neq \thermalDisplacement(t)$, but the following relations hold:
\begin{subequations}
\begin{align}
    \displacementSpring(\zeit) &= \displacement_\text{th}(\zeit) \qquad \rightarrow \qquad \posTS(\zeit) = 0 \eqCom \\
    \displacementSpring(\zeit) &< \displacement_\text{th}(\zeit) \qquad \rightarrow \qquad \posTS(\zeit) > 0 \eqDot
\end{align}
\end{subequations}
The case where $\displacementSpring>\displacement_\text{th}$ is impossible because, in this case, the electrodes would no longer touch the sheet. Consequently, $\posTS$ equals the thermal expansion minus the control-induced adjustment:
\begin{equation}
    \posTS(\zeit) = \displacement_\text{th}(\zeit) - \displacementSpring(\zeit) \eqDot
\end{equation}

\begin{figure}[]
    \centering
    \includegraphics[width=0.9\textwidth]{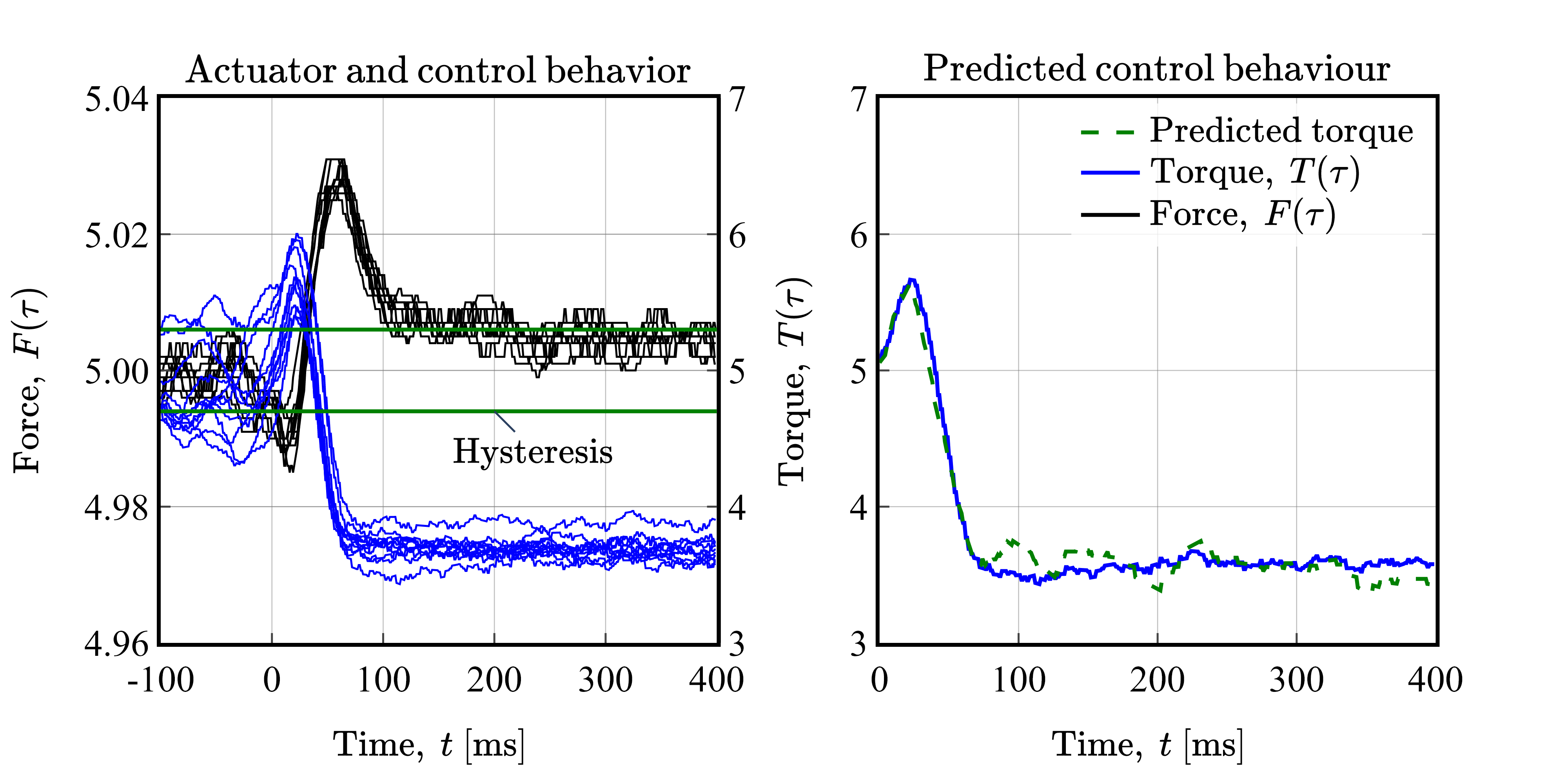}
    \caption{Left: Interplay between the control signal and the actuator. The evident hysteresis band is overlaid. Right: Estimated PID control behavior by fitting the PID constants.}
    \label{fig:pid_force_hysteresis}
\end{figure}

We estimate the step response of the welding control by assuming a \acrfull{PID} control. The control maintains an approximately constant displacement at equilibrium; deviations from the set force value are countered by dynamically adjusting the torque $T$ of the electric motor driving the upper electrode. This interplay is evident in \fig~\ref{fig:pid_force_hysteresis}. During the onset of the welding current, $\thermalDisplacement$ exerts a force against the electrodes, yielding an error $\error(t)$ relative to the set value. The error $\error(t)$ equals the set displacement $\displacement_0$ minus the actual displacement:
\begin{equation}
    \error(t) = \displacement_0 - \posTS(t) \eqDot
\end{equation}
Because the reference displacement $\displacement_0$ is set to zero, the error equals $\error = - \displacement(\zeit)$. 
%
%
A generic \acrshort{PID} control discretized at each control time step $\tau$ can describe the response:
\begin{equation}
    \displacementSpring(\tau) = \text{PID} \left( \error, \propConst, \inteConst, \diffConst \right) = \propConst \error(\tau) + \inteConst \sum_{i=0}^{\tau} \error(i) \Delta \zeit + \diffConst \frac{e(\tau) - \error(\tau - 1)}{\Delta \zeit} \eqDot \label{eq:PID_response}
\end{equation}
The previous error $e(\tau - 1)$ can be readily computed from the most recent time step, and $\Delta \zeit$ is one millisecond. The integral component can be restricted to a fixed number of previous steps to limit computation, where only the recent errors are considered in the accumulation term.
We estimate the \acrshort{PID} constants by minimizing the \acrlong{MSE} between the actuator, the torque $T(\tau)$, and the PID response, given the error $\error(\tau) = F_0 - \springConst \displacementSpring(\tau)$. The optimization is defined to yield the estimated proportional, integral and derivative constant:
\begin{equation}
    \hat{K}_\text{p}, \hat{K}_\text{i}, \hat{K}_\text{d} = \argmin_{\propConst, \inteConst, \diffConst} \left\{ \acrshort{MSE}\left( T, \text{PID} \left(\error, \propConst, \inteConst, \diffConst \right) \right) \right\} \eqDot \label{eq:argmin_torque_force}
\end{equation}
Based on this optimization, the resulting estimated PID constants for the controller are: $\hat{K}_\text{p}=5.3 \ [-]$, $\hat{K}_\text{i}=0.2 \ [\text{s}^{-1}]$ and $\hat{K}_\text{d}=0.003 \ [\text{s}]$.
Hysteresis in the force control is observable within a range of approximately 0.1\perc, corresponding to a tolerance band of $\pm 5$ newton around a 5 kN setpoint. This observation is evident in \fig~\ref{fig:pid_force_hysteresis} and manifests in the signal as values slightly below the setpoint before the onset of welding and marginally above it after the control response. 
\section{Methods}
This chapter details the methods for modeling aluminum RSW and integrating experimental data. We begin by outlining the problem setup and the governing equations, including the key assumptions and simplifications made. Special attention is given to the electric contact heat generation at the faying surface.
After, we review how the introduced equations and RSW-specific considerations are implemented as loss functions in a PINN.
A key focus is on integrating experimental data directly into the PINN's training. This integration is achieved through a dynamic temperature-dependent material parameter update procedure and a progressive experimental data assimilation approach that employs a sigmoid-shaped fading-in function. Furthermore, a custom rolling window scheduler and early stopping are implemented to ensure network convergence despite the introduction of new optimization objectives.

\subsection{Problem setup and constitutive equations}
\Acrlong{Al} \acrshort{RSW} represents a complex multiphysics problem involving coupled mechanical, thermal, and electrical fields.
As an industrial welding process, the primary focus is on the temperature field because it determines the liquidus temperature and, thereby, the joining area of the metal sheets. The key characteristic of \acrshort{RSW} is that the heat is not supplied externally but is generated internally through electrical and mechanical interactions. Magnetic and chemical effects are typically negligible in this context \autocite{Li2009,Wei1990}, albeit electromagnetic effects have been shown to influence steel welding \autocite{Wei2014,Li2007}. The impacts of these effects are negligible for \acrlong{Al}, as discussed by \textcite{Yang2004}. To simplify the analysis and considering the computational constraints when using PINNs, we focus solely on the temperature and displacement fields, and the influence of \mbox{phase-change-related} latent heat is not considered because enthalpy is negligible for the transition from solid to liquid. 
Contact is not calculated explicitly, and sticking is assumed. To make the computational analysis feasible, we simplify the model into two models, a 1D model for the development of our methods and a \acrlong{2D} rotationally symmetric model.
Due to the high computational cost of repeated runs in 2D models, initial testing was done on 1D problems. One dimension also reduces model complexity and computational cost, which is desirable for industrial applications. The approach was then applied to an axially symmetric 2D model to represent the welding process more realistically.
One dimension reduces model complexity and computational cost, which is desirable for industrial applications.
However, the gold standard quality metric is the nugget diameter, which lies within the radial axis. Including the nugget diameter necessitates a transition to a 2D model.

We propose a model that simplifies the industrial machine setup, assuming a substitute model can approximate the system to a sufficient degree. This configuration and the sensor's location are illustrated in \fig~\ref{fig:ersatzsystem}.
The system is modeled using a substitute in which the \cShaped\ welding arm is replaced by a spring. The replacement spring constant for the static electrode has been calculated based on linear elasticity to have a stiffness of $\springConst = 6,666 \ [\springConstUnit]$. The deformation of the electrode arm is assumed to be elastic and corresponds linearly to the spring deformation, a point that is elaborated with the control equation. We also assume that there is no significant thermal expansion within the welding gun due to cooling of the electrode caps. This is because it is subjected to a limited number of welds, \ie\ up to 30 welds. The substitute system also assumes that the angular momentum and body force acting on the welding gun can be neglected, as they have been calculated to have a negligible impact. Additionally, the system is considered \quasiStatic\ with a set electrode force $\forceMax$ and contact area $\apparentArea$ at the onset of electric current flow.

In our approach, the equations are the heat conduction \autocite{Lienhard2019} and the equation of linear elasticity \autocite{Landau2005}. We define spatial coordinates as $\textbf{x} = (x,y,z)$. We assume static equilibrium without body force for an isotropic, homogeneous material for temperature $\temperature$ and stress $\stressField$:
\begin{align}
    \density(T) \specHeatCap(T) \pdv{\temperature(\textbf{x},t)}{\zeit} &= \nabla \cdot \left( \thermCond(T) \nabla \temperature(\textbf{x},t) \right) + \specElectrRes(T) \currentDensity(t)^2 \eqCom \label{eq:temperature_governing_equation} \\
    \nabla \cdot \stressField (\textbf{x},t) &= 0 \eqDot \label{eq:displacement_governing_equation}
\end{align}
Here, $\density$ denotes density, $\specHeatCap$ is the specific heat capacity, $\thermCond$ is the thermal conductivity, and $\specElectrRes$ electric resistivity.
Typical values are listed in \tab~\ref{tab:typical_values}. The constitutive relations for linear elasticity are:
\begin{align}
\begin{split}
    \stressField &= \firstLameConst(T) \left( \nabla \cdot \displacementField(\textbf{x}, t) - 3 \thermExp(T) \temperature(\textbf{x},t) \right) \identity + 2 \secondLameConst(T) ( \strainField - \thermExp(T) \temperature(\textbf{x},t) \identity) \eqCom \\
    \strainField &= \frac{1}{2} \left( \nabla \displacementField(\textbf{x}, t) + (\nabla \displacementField(\textbf{x}, t))^\intercal \right) \eqCom \qquad
    \firstLameConst(T) = \frac{\eModul(T) \poissonRatio}{(1 + \poissonRatio)((1 - 2 \poissonRatio)} \eqCom \qquad
    \secondLameConst(T) = \frac{\eModul(T)}{2(1 + \poissonRatio)} \eqDot
\end{split}
\end{align}
Here, $\displacementField(\textbf{x}, t)$ denotes displacement, $\thermExp(T)$ is the coefficient of linear thermal expansion, $\firstLameConst(T)$ and $\secondLameConst(T)$ are the first and second \lame\ constants, $\strainField$ is the strain, $\eModul(T)$ is \young's modulus, and $\poissonRatio$ is \poisson\ ratio. For brevity, the dependence on $T$ and $\textbf{x}$ is hereafter implicit. 
Assuming radial symmetry, the governing equations, \equ~\ref{eq:temperature_governing_equation} and \equ~\ref{eq:displacement_governing_equation}, can be formulated in cylindrical coordinates. The temperature field is therefore $\temperature(r, z, \zeit)$. We assume that the displacement field is totally clamped in the radial direction in our application due to the surrounding material. The assumption $\displacement_r = 0$ is physically plausible if the system is perfectly clamped in the radial direction, yielding a dependence only on $z$ with $u(z,t)$. The rotationally symmetric \acrshort{2D} form reads:
\begin{align}
    \density \specHeatCap \pdv{\temperature}{\zeit} &= \frac{1}{r} \pdv{}{r} \left( \thermCond r \pdv{\temperature}{r} \right) + \pdv{}{z} \left( \thermCond \pdv{\temperature}{z} \right) + \specElectrRes \currentDensity^2 \eqCom \label{eq:temperature_governing_equation_2d} \\
    (\firstLameConst + 2 \secondLameConst) \frac{\partial^2 u}{\partial z^2} &= (3 \firstLameConst + 2 \secondLameConst) \thermExp \pdv{\temperature}{z} \eqDot \label{eq:displacement_governing_equation_2d}
\end{align}
%

\begin{figure}
    \centering
    \includegraphics[width=\textwidth]{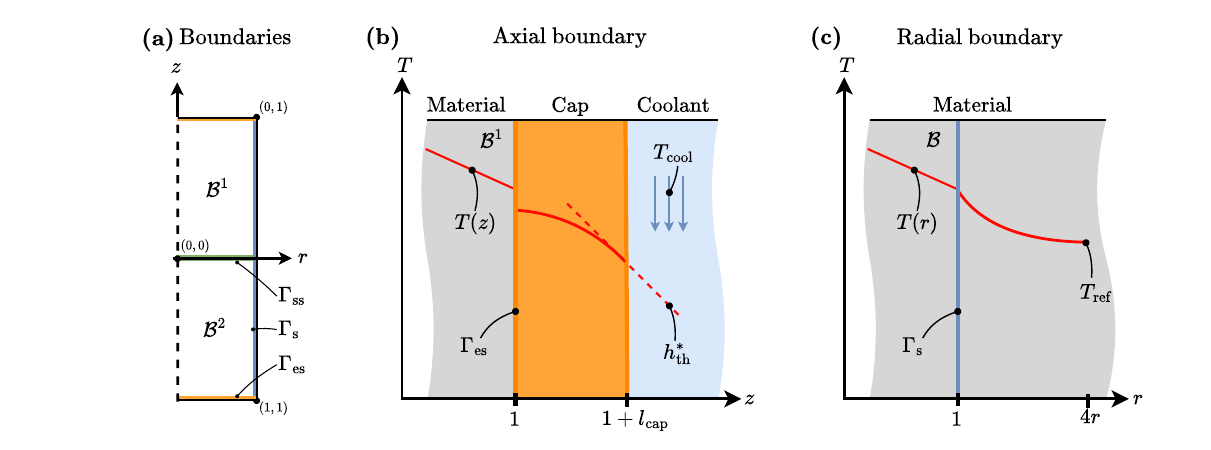}
    \caption{Left: The configuration of the upper $\body^1$ and lower sheet $\body^2$. Middle: An illustration following \textcite{Budynas2020} of the \neumann\ \acrlong{BC} at the upper sheet and the electrode cap, along with the effect of electrode cooling. The heat transfer coefficient $h_\text{th}^*$ describes the flux between the electrode cap and the coolant. Right: An illustration of the radial boundary with a reference temperature $\refTemp$.}
    \label{fig:configuration_neumann}
\end{figure}

%
%
The initial condition for $\temperature$ is set to a reference room temperature $\refTemp$ and $\displacement=0$. The thermal boundary condition on the \electrodeSheet\ interface is defined with a heat flux. The boundaries are visualized in \fig~\ref{fig:configuration_neumann}, with $\boundary = \electrodeSheetBC \cup \sheetSheetBC \cup \sheetBC$. Here, $\electrodeSheetBC$ is the boundary towards the electrode caps, $\sheetSheetBC$ is the faying surface, and $\sheetBC$ is the radial boundary towards the sheet material. Using the \biot\ number $\biotNumber = \heatTransCoeff \thermCond^{-1}$, the axial boundary is defined:
\begin{equation}
    \pdv{\temperature}{z} = \biotNumber \left[ \temperature(0, \zeit) - \temperature_\text{ref} \right] \qquad \text{on} \qquad\electrodeSheetBC \eqDot
    \label{eq:neumann_BC}
\end{equation}
The heat transfer coefficient $\heatTransCoeff$ has been measured by \textcite{Piott2020a} to be $25,000 \ [\heatTransCoeffUnit]$ at a coolant flow rate of 3 liters per minute and a comparable machine setup.
Cylindrical coordinates require a different approach for the radial heat flux \autocite{Lienhard2019} than the axial heat flux. As heat flows to a reference temperature, the cylindrical geometry changes, illustrated in \fig~\ref{fig:configuration_neumann}c. The condition reads at $r=1$:
\begin{align}
    \pdv{\temperature}{r} &= \frac{1}{\ln L} \left[ \temperature_\text{ref} -\temperature(r=1) \right] \qquad \text{on} \qquad \sheetBC \eqDot
    \label{eq:radial_heat_flux}
\end{align}
Here, $L$ is a chosen distance at which $\temperatureDim_\text{ref}$ is known. From experiments, we observed that at a distance of $L = 20$ millimeters, the temperature of the sheet material is unaffected. 
\subsection{Electric contact heat generation at the faying surface}
\begin{figure}
    \centering
    \includegraphics[width=0.95\textwidth]{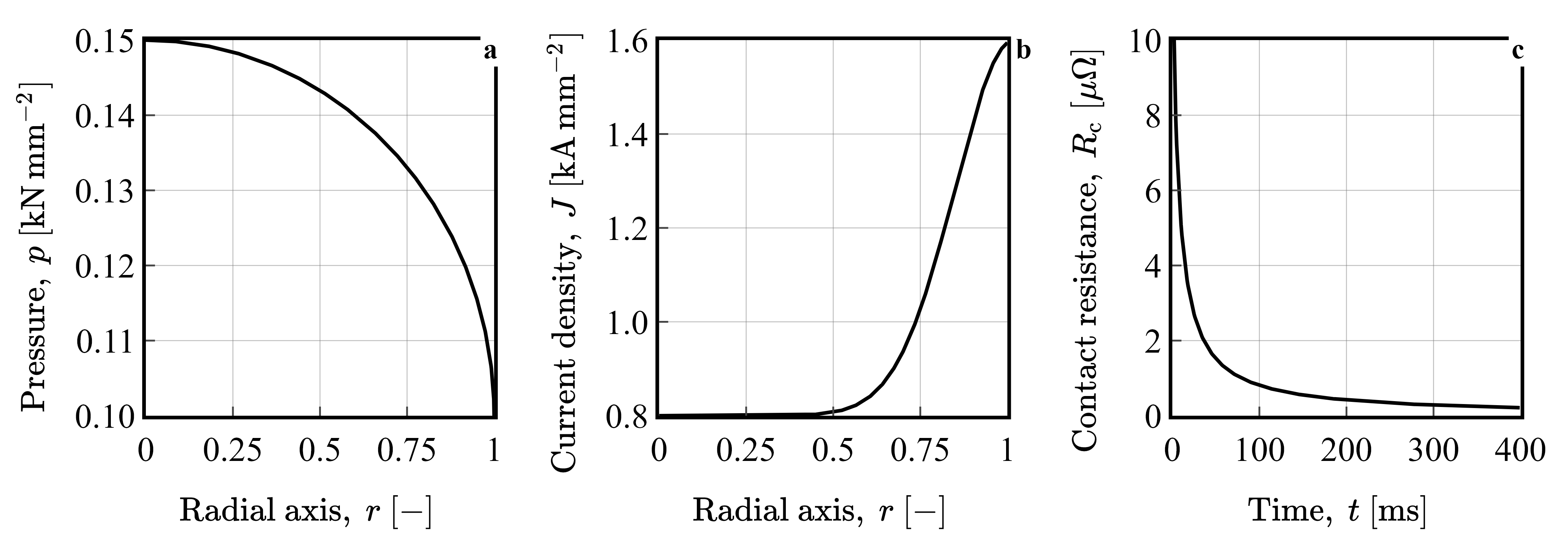}
    \caption{\hertzian\ contact pressure, \textcite{Greenwood1958} current density, and \textcite{Kaars2016a} logarithmic decay of $\contactRes$, all on $\sheetSheetBC$.}
    \label{fig:boundary_conditions}
\end{figure}

Modeling the contact behaviour on a microscopic level is considered a complicated process \autocite{Wei1990,Eagar1990} because the initial electrical and thermal contact situation is hard to measure. 
Quantitatively, \textcite{Kaars2016b} proposed a \temperatureDependent\ exponentially decaying contact resistance $\contactRes$ between two sheets of \mbox{hot-dipped} steel. An exemplary decay is shown in \fig~\ref{fig:boundary_conditions}c. \textcite{Wang2001} also studied the contact resistance for steel \acrshort{RSW} and found that the initial rapid decrease of the dynamic electrical resistance is primarily due to the reduction in film resistance $\filmRes$.
Two key electrical contact resistances arise in RSW: one at the \electrodeSheet\ interface and another at the faying surface. The heat generated here is of particular interest, as the remains of the adhesive create a high contact resistance, and the region experiences the highest heat accumulation. This is because the faying surface is farthest from the cooled electrodes. In contrast, the heat generated at the cooled electrode-sheet interface is minimal.
The contact resistance $\contactRes$ consists of constriction $\constrRes$ and film resistance $\filmRes$. The former is primarily caused by surface roughness, which constricts current streamlines at discrete contact points, and the latter by the presence of films of matter, such as oil, adhesives, or other contaminants, commonly found in industrial settings.
The resistances for a local contact spot have been derived by \textcite{Holm1967} as:
\begin{equation}
    \constrRes = \frac{\specElectrRes}{2 a} \eqCom \quad \text{and} \quad \filmRes = \frac{\specElectrRes^f \filmThk}{a^2 \pi} \eqCom \label{eq:holm_constriction_film}
\end{equation}
where $a$ is the effective radius, $\specElectrRes$ is the electric resistivity of the material or film, and $\filmThk$ is the film thickness. The effective radius is used because the apparent area $\apparentArea$ differs from the real area $ \realArea$ \autocite{Bowden2001}, sometimes termed the \mbox{load-bearing} area.
The electrical resistance of a \mbox{metal-to-metal} contact can be assumed to consist of uniformly distributed contact spots $a_i$ \autocite{Greenwood1958}:
\begin{equation}
    \realArea = \sum_{i=1}^n a_i^2 \pi = n a^2 \pi \eqDot
\end{equation}
Based on experimental evidence, \textcite{Thomas1980} demonstrated that the real contact area shows a linear dependence on the applied load, that is, $\forceMax$ and the \vickers\ hardness $\hardness$:
\begin{equation}
    n \pi a^2 = \frac{\forceMax}{\hardness} \eqDot \label{eq:woo_thomas}
\end{equation}
Holm and Greenwood referenced \acrshort{RSW} as an application example, and research on it has since been conducted. 
For example, \textcite{Crinon1998} experiment on pretreated surfaces and \textcite{James1997} conduct a rudimentary \acrshort{FEM} analysis of the contact resistance using Holm's formula. 
A model for $\contactRes$ has been developed by \textcite{Wang2001} with a multiphysics setup. The authors stress that the mechanical and electrical properties of the materials, such as hardness and resistivity, are temperature-dependent, and they applied this method in subsequent studies \textcite{Wei2012,Wei2014}. Their definition of local contact resistance builds upon \equ~\ref{eq:holm_constriction_film} and \equ~\ref{eq:woo_thomas} and is given by:
\begin{equation}
    \contactRes^A = \frac{\specElectrRes}{2} \sqrt{\frac{n \pi \hardness}{\forceMax}} + \specElectrRes^f \frac{n \filmThk \hardness}{\forceMax} \eqDot \label{eq:wei_contact_resistance}
\end{equation}
Here, $n$ is the number of contact spots, $\specElectrRes^\text{f}$ is the electric resistivity of the film, and $\filmThk$ is the film thickness. For epoxy resins, like the weld bonding adhesive, a typical value for $\specElectrRes^\text{f}$ is between $10^6$ and $10^9$ \mbox{ohm-meters} \autocite{Tang2013,Mousavi2022}. The \acrlong{RHS} of \equ~\ref{eq:wei_contact_resistance} corresponds to the constriction and film resistance, respectively. It is commonly argued that during welding, the number of contact spots remains constant while their size increases. Directly measuring the real contact area is nearly impossible, especially in industrial settings.

Alternatively, a model can be defined with the essential relation described by \equ~\ref{eq:woo_thomas}, directly relating the contact resistance to the applied force and material hardness.
\textcite{Weissenfels2009} considered spreading resistance \textcite{Holm1967} based on a projection of 3D models on the surface and considering the parallel connection of the spots at the contact surface. The authors defined a pressure-dependent contact resistance for electric contact between two metals. To appropriately compare both models, we extend the \mbox{constriction-based} contact resistance by a film resistance term for the forward model:
\begin{equation}
    \contactRes^B = \frac{1.05 \specElectrRes}{4} \sqrt{\frac{\pi \hardness}{\forceMax}} + \specElectrRes^f \frac{\filmThk \hardness}{\forceMax} \eqDot
    \label{eq:weissenfels}
\end{equation}
%

%
%
The faying surface $\sheetSheetBC$ is of key interest at which the weld nugget develops. We treat the thermal heat generated on $\sheetSheetBC$ as an additive heat source to the heat equation, \cf\ \equ~\ref{eq:temperature_governing_equation}.
Although \textcite{Chen2019} suggested that most adhesive is squeezed out during $\zeitS$, \textcite{Zhao2018} demonstrated an increased $\contactRes$ due to the remaining adhesive. The authors investigated the effect of the adhesive using \mbox{dual-phase} steel and summarized that the nugget diameter is increased by higher $\contactRes$ with weld bonding \autocite{Kaars2016a,Piott2019,Zhao2018}.
Assuming $n$ parallel contact spots, the contact heat is defined as the sum of bulk heat generation and contact heat defined in \equ~\ref{eq:wei_contact_resistance} and using $\currentDensity$ defined in \equ~\ref{eq:material_parameter}:
\begin{align}
    q_c^A &= \specElectrRes \currentDensity^2 + \frac{\apparentArea}{\filmThk} \frac{\contactRes^A}{n} \currentDensity^2 \eqCom \qquad \text{at} \qquad z=0.5 \eqCom
    \label{eq:contact_res_heat} \\
    q_c^B &= \specElectrRes \currentDensity^2 + \frac{\apparentArea}{\filmThk} \contactRes^B \currentDensity^2 \eqCom \qquad \text{at} \qquad z=0.5 \eqDot
    \label{eq:contact_res_heat_B}
\end{align}

Pressure and current density must be considered radially when expanding the setup with a radial dimension. The applied local pressure is assumed to follow a \hertzian\ pressure distribution, a model with a \mbox{well-established} analytical solution, as shown in \fig~\ref{fig:boundary_conditions}a.
Concerning the contact heat generation, theoretical discussions by \textcite{Greenwood1958} suggested that the temperature distribution relies solely on the current density $\currentDensity$ and the contact geometry. It was proposed that $\temperature$ follows approximately a normal distribution increasing towards the periphery of the contact area, as shown in \fig~\ref{fig:boundary_conditions}b.
\subsection{Reformulation and non-dimensionalization}
We introduce two key variables of the heat equation: thermal diffusivity $\thermDiffusivity$ and the ratio of $\specElectrRes$ to $\density$ and $\specHeatCap$, which is termed electro-thermal dissipation $\electrThermDissipation$:
\begin{equation}
    \thermDiffusivity = \frac{\thermCond}{\density \specHeatCap} \eqCom \qquad \text{and} \qquad \electrThermDissipation = \frac{\specElectrRes}{\density \specHeatCap} \eqDot
    \label{eq:reformulation}
\end{equation}
\NonDimensionalization\ is a common method in mathematics and physics to transform a given system into a dimensionless form \autocite{Wang2023Expert,Langtangen2016}. 
In the case of \acrshort{PINN}s, it acts as normalization of the data, which is often implemented to enhance numerical optimization and convergence of \mbox{gradient-based} learning \autocite{Jung2022}. For this reason, \nonDimensionalization\ is important for \acrshort{PINN}s, among others, because it alleviates convergence issues \autocite{Wang2021,Wang2022}.
We non-dimensionalize \equ~\ref{eq:temperature_governing_equation_2d} and \equ~\ref{eq:displacement_governing_equation_2d} by introducing the following characteristic dimensionless variables:
For clarity, in the subsequent discussion, we omit the explicit notation of the dimensionless variables:
\begin{equation}
    \rDim = \frac{r}{\rChar}\eqCom \quad \zDim = \frac{z}{\zChar}\eqCom \quad \zeitDim = \frac{\zeit}{\zeitChar} \eqCom \quad \temperatureDim = \frac{\temperature - \temperature_0} {\temperatureChar} \eqCom \quad \displacementDim = \frac{\displacement}{\displacementChar} \eqDot
\end{equation}
Here, the variables $\rChar$, $\zChar$, $\zeitChar$, $\temperatureChar$, and $\displacementChar$ represent the corresponding characteristic scales for the radial, axial, temporal, temperature, and displacement variables.
We define a characteristic time scale to consider the primary effect in the axial direction. Defining the characteristic displacement as:
\begin{equation}
    \zeitChar = \frac{\densityChar \specHeatCapChar \zChar^2}{\thermCondChar} \eqCom \qquad \text{and} \qquad \displacementChar = \frac{(3 \firstLameConst + 2 \secondLameConst) \temperatureChar \zChar \thermExp}{(\firstLameConst + 2 \secondLameConst)} \eqDot
    \label{eq:time_scale_def}
\end{equation}
After grouping, the non-dimensionalized form reads:
\begin{align}
    \pdv{\temperatureDim}{\zeitDim} &= \frac{\zChar^2}{\rChar^2} \frac{1}{\rDim} \pdv{}{\rDim} \left( \thermCond \rDim \pdv{\temperatureDim}{\rDim} \right) + \pdv{}{\zDim} \left( \thermCond \pdv{\temperatureDim}{\zDim} \right) + \frac{\zChar^2}{\temperatureChar} \electrThermDissipation \currentDensity^2 \eqCom \\
    \frac{\partial^2 \displacementDim_z}{\partial \zDim^2} &= \pdv{\temperatureDim}{\zDim} \eqDot
    \label{eq:nondimensionalized_PDE}
\end{align}
%
%
\begin{table}
  \centering
  \begin{minipage}[t]{0.45\textwidth}
    \vspace{0pt}
    \centering
    \begin{tabular}{lccc}
    \toprule
    \textbf{Typical values}  & \textbf{Unit} & \textbf{Value} \\
    \toprule
    $\specHeatCap$ & $\specHeatCapUnit$ & $890$ \\
    $\density$ & $\densityUnitMM$ & $2.7 \times 10^{-6}$ \\
    $\thermDiffusivity$ & $\thermDiffusivityUnitMMMS$ & $0.06$ \\
    $\thermCond$ & $\thermCondUnitMM$ & $0.16$ \\
    $\specElectrRes$ & $\specElectrResUnitMM$ & $3.66 \times 10^{-5}$ \\
    $\filmThk$ & mm & $10^{-5}$ \\
    $\specElectrRes^f$ & $\specElectrResUnitMM$ & $10^{5}$ \\
    $\poissonRatio$ & $\poissonRatioUnit$ & 0.33 \\
    $\hardness$, $\eModul$ & $\text{N}\,\text{mm}^{-2}$ & $430$, $70$ \\
    $n$ & $-$ & 120,000 \\
    $\apparentArea$ & $\text{mm}^2$ & 79 \\
    $\refTemp, \liqTemp, \solTemp$ & $\temperatureUnit$ & 20, 660, 630 \\
    \bottomrule
\end{tabular}
\caption{Typical values for \acrlong{AL6}.}
\label{tab:typical_values}
  \end{minipage}\hfill
  \begin{minipage}[t]{0.45\textwidth}
    \vspace{0pt}
    \centering
    \begin{tabular}{lccc}
    \toprule
    \textbf{Characteristic value}  & \textbf{Unit} & \textbf{Value} \\
    \toprule
    $r_\text{c}$ & $\text{mm}$ & 5 \\
    $\zChar$ & mm & 3 \\
    $\zeitChar$ & ms & $\densityChar \specHeatCapChar \zChar^2 \thermCondChar^{-1}$ \\
    $\temperatureChar$ & $\temperatureUnit$ & $\liqTemp$ \\
    $\displacementChar$ & mm & $\temperatureChar \thermExp \zChar$ \\
    \bottomrule
\end{tabular}
\caption{Characteristic values for the \nonDimensionalization.}
\label{tab:characteristic_values}
  \end{minipage}
\end{table}
\subsection{Physics-informed neural network setup}
\begin{figure}
    \hspace{1.2cm}
    \includegraphics[width=\textwidth]{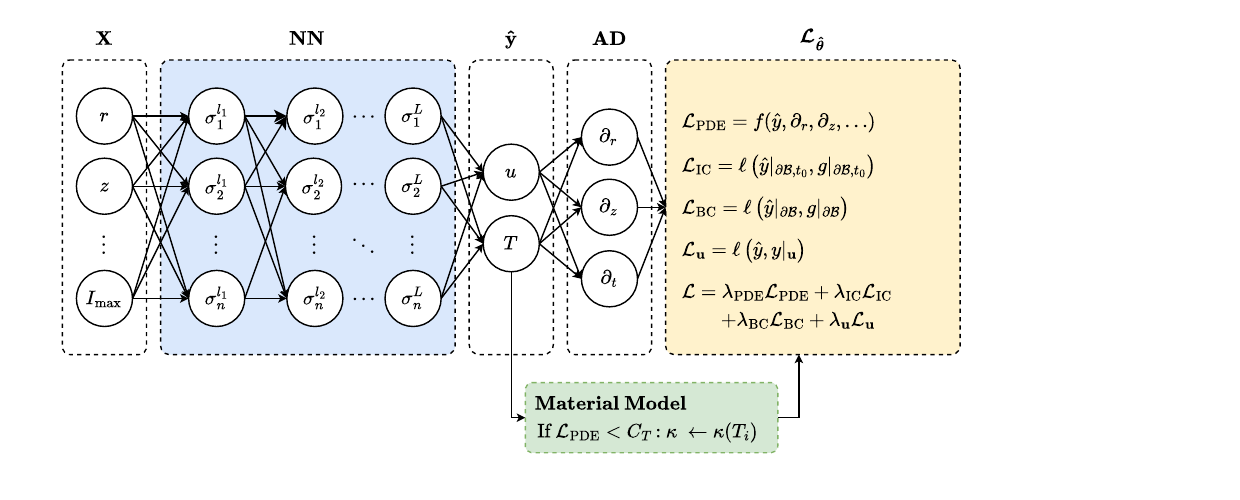}
    \caption{Illustration of the \acrshort{PINN} setup for \acrshort{RSW} with \temperatureDependent\ material model and dynamic displacement.}
    \label{fig:PINN_RSW_material_model}
\end{figure}

A general formulation of \acrshort{PINN}s \autocite{Raissi2017a,Raissi2017b} starts with a \acrfull{NN} $f_\text{\acrshort{NN}}$ defined with a set of internal, learnable parameters, \ie\ weights and biases denoted as $\params$, and configurations chosen before training that steer the learning process, \ie\ hyperparameters $\hyperParam$. Within a single layer $\layer \in L$, the weight matrix and bias vector are referred to as $\weights^{\layer} \in \mathbb{R}^{z_{\layer} \times z_{\layer - 1}}$ and $\biases^{\layer} \in \mathbb{R}^{z_{\layer}}$ \autocite{Lu2020}. with this setup, a \acrlong{NN} can be defined recursively with a neuron $z$:
\begin{equation}
\begin{array}{rl}
    \text{input layer:}  & z^0(\predictor) = \predictor \in \mathbb{R}^{d_\text{in}} \eqCom \\
    \text{hidden layer:} & z^\layer(\predictor) = \activationFunc^\layer \left( \weights^l z^{\layer-1}(\predictor) + \bias^\layer \right) \in \mathbb{R}^{n_\layer} \eqCom \quad \forall \layer \in [1, \layers - 1] \eqCom \\
    \text{output layer:} & z^\layers(\predictor) = \weights^\layers z^{\layers - 1}(\predictor) + \bias^\layers \in \mathbb{R}^{d_\text{out}} \eqDot
    \label{eq:neural_network}
\end{array}
\end{equation}
%
The \continuousTime\ approach uses a $\fNN$ to approximate the hidden solution, parametrizing and training to learn $\params$, thereby approximating the \acrshort{PDE}. The rationale behind this approach lies in utilizing the universal approximation theorem, asserting that $\fNN$ can effectively approximate any function \autocite{Cuomo2022}, and also coupled fields:
\begin{equation}
    \fNN (r, z, \zeit; \params) \approx \left[ \temperatureEst (r, z, \zeit), \displacementEst(r, z, \zeit) \right] \eqDot
\end{equation}
\fig~\ref{fig:PINN_RSW_material_model} illustrates the \acrshort{PINN} setup, where $r$, $z$, and $\zeit$ represent the space and time domains.
%
%
The $\fNN$ is regularized with the partial differential equations for both fields, \cf\ \equ~\ref{eq:nondimensionalized_PDE}.
Incorporating a welding schedule into a \acrshort{PINN} requires \mbox{time-dependent} heat generation. A realistic welding schedule introduces an upslope and hold time, as defined in \equ~\ref{eq:stromprofil}. However, this is nontrivial due to the discontinuities in the first derivative of the welding schedule. For the temperature field, the non-dimensionalized heat equation is set as the PDE loss, denoted as $\lossPDE^T$.
The initial condition is enforced with $\lossIC^T$, where the initial temperature is referenced to room temperature, which is set to 23 degrees, $\temperature = \temperature^\prime - \roomTemp$.
The boundary consists of axial and radial heat flux and is regularized by $\lossBC^T$. The axial heat flux, defined in \equ~\ref{eq:neumann_BC}, can be directly implemented as a boundary loss function. Because the direction of the heat flux is unknown to the \acrlong{AD}, the absolute value of the temperature gradient is necessary to enforce an outward flux.
The radial heat flux is defined with \equ~\ref{eq:radial_heat_flux}:
\begin{subequations}
\begin{align}
    \lossPDE^T &= \frac{1}{| \datasetPDE |} \sum_{i=1}^{| \datasetPDE |} \left[ \pdv{\temperatureEst_i}{\zeit} - \frac{\zChar^2}{\rChar^2} \frac{1}{r} \pdv{}{r} \left( \thermCond r \pdv{\temperatureEst_i}{r} \right) - \pdv{}{\zDim} \left( \thermCond \pdv{\temperatureEst_i}{z} \right) - Q \right]^2 \label{eq:heat_eq_loss} \eqCom \\
    \lossIC^T &= \frac{1}{| \datasetIC |} \sum_{i=1}^{| \datasetIC |} \left[ \temperatureEst_i - \temperature_\text{ref} \right]^2 \eqCom \\
    \lossBC^T &= \frac{1}{| \datasetBC |} \sum_{i=1}^{| \datasetBC |} \left[ | \underbrace{\pdv{\temperatureEst_i}{z} | - \biotNumber \left[ \temperature(r,0,\zeit) - \temperature_\text{ref} \right]}_\text{Axial flux} \right]^2 + \left[ | \underbrace{\pdv{\temperatureEst_i}{r} | - \frac{1}{\ln L} \left[ \temperature_\text{ref} -\temperature(1,z,t) \right]}_\text{Radial flux} \right]^2 \eqDot \label{eq:neumann_loss}
\end{align}
\end{subequations}
Here, $\temperatureEst_i$ denotes the i-th estimate, and $\refTemp$ has different reference values in each term. The heat source depends on the bulk material or the faying surface, as defined in \equ~\ref{eq:contact_res_heat}:
\begin{equation}
    Q = \left\{
    \begin{array}{cl}
        \frac{\zChar^2}{\temperatureChar} \electrThermDissipation \currentDensity^2  & : \ r_i, z_i \hspace{2mm} \in \quad \body \ \\
        \frac{\zChar^2}{\temperatureChar} \frac{1}{\specHeatCap \density} q_c^{A/B} & : \ r_i, z_i \quad \text{on} \quad \sheetSheetBC \label{eq:explicit_PDE_loss}
    \end{array} \right.
\end{equation}
The number of contact spots in \equ~\ref{eq:wei_contact_resistance} is set as a learnable parameter for the \acrlong{NN} to steer contact heat generation.
The available data for \acrlong{Al} \acrshort{RSW} includes the measurement of the formed nugget diameter $\nuggetDia$. The measurement of $\nuggetDia$ is used to penalize the \acrshort{PINN} to a temperature above the liquidus at $\zeit = 560$. Afterward, the cooling period begins. At this point, the maximum temperature is reached. This information is referred to as a goal loss $\lossGoal$, as defined in \autocite{Seo2024}. This penalty can be added to find a solution that not only solves the governing equations but also achieves an additional objective. 
In $\dataset_{\nuggetDia}$, points are identified where the temperature must have surpassed the liquidus value to form the molten nugget. The proposed loss formulation is based on two key considerations. First, it avoids penalizing temperatures that exceed the liquidus temperature, focusing only on the available information: the temperature is above this threshold rather than specifying an exact value. Second, it preserves the computational graph that the \acrlong{AD} generates. This is accomplished by applying a masking operation to $\temperatureEst$ during backpropagation and computing the temperature difference relative to $\liqTemp$. We define $\dataset_{\nuggetDia}$ on $\sheetSheetBC$ and \mbox{non-dimensionaize} the temperature using $\temperatureChar$:
\begin{equation}
    \loss_{\nuggetDia} = \frac{1}{| \dataset_{\nuggetDia} |} \sum_{i=1}^{| \dataset_{\nuggetDia} |} \left[ \min \left\{ \temperatureEst(r,0.5,t), \, \liqTemp \right\} - \liqTemp \right]^2 \eqDot \label{eq:goal_loss_dp}
\end{equation}
%

\begin{figure}
    \hspace{1.8cm}
    \includegraphics[width=0.8\textwidth]{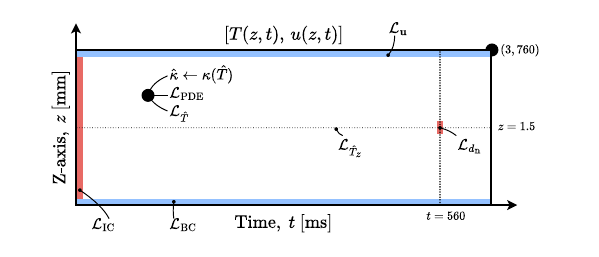}
    \caption{Illustration of the losses for \acrshort{RSW}. Blue indicates the \acrlong{BC} and red the \acrlong{IC} and $\loss_{\nuggetDia}$. The goal losses $\loss_\posTS$ are indicated in yellow, where the two lines are illustrative for $\dataset_\posTS$ and $\dataset_\resTS$, respectively. The dotted line represents the location of $\lossContinuity$, while one collocation point demonstrates the application of the material update, $\lossPDE$ and $\lossTemp$. The \raxis\ is omitted for this illustration.
    }
    \label{fig:goal_losses}
\end{figure}

%
%
For the displacement field, the network regularizes the governing equation with the loss $\lossPDE^\displacement$. The initial condition is enforced to be zero with $\lossIC^\displacement$.
The radial displacement is constrained to zero $\lossBC^\displacement$. 
The experimental data is added using $\loss_\posTS$ at $z=1$. It is the estimated welding control response $\displacementSpring$, defined in \equ~\ref{eq:PID_response}, minus the experimental dynamic displacement $\displacement(\zeit)$. The loss functions for the displacement field are:
\begin{subequations}
\begin{align}
    \lossPDE^\displacement &= \frac{1}{| \datasetPDE |} \sum_{i=1}^{| \datasetPDE |} \left[ \frac{\partial^2 \displacementEst_i}{\partial z^2} - \pdv{\temperatureEst_i}{z} \right]^2 \eqCom \\
    \lossIC^\displacement &= \frac{1}{| \datasetIC |} \sum_{i=1}^{| \datasetIC |} \displacementEst_i(z,0)^2 \eqCom \\
    \lossBC^\displacement &= \frac{1}{| \datasetBC |} \sum_{i=1}^{| \datasetBC |} \displacementEst_i(0, \zeit)^2 \quad \text{and} \quad
    \loss_\posTS = \frac{1}{| \datasetBC |} \sum_{i=1}^{| \datasetBC |} \left[ \displacementEst(1, \zeit) - \displacementSpring(\zeit) - \posTS(t) \right]^2 \eqDot 
\end{align}
\end{subequations}
%
%
%
Radial symmetry may be defined at $r=0$ as an equilibrium condition. The same condition can be defined for the \zaxis\ at $z=0.5$:
\begin{equation}
    \pdv{\temperature}{r} = 0 \quad \text{on} \quad r=0 \eqCom \qquad \text{and} \qquad \pdv{\temperature}{z} = 0  \quad \text{on} \quad \sheetSheetBC \eqDot
\end{equation}
The latter condition is not as trivial as the radial equilibrium. If the material sheet thickness is unequal, the \sheetSheet\ interface is no longer at $z=0.5$, and we can enforce it at another $z$ value. 
The governing equation naturally constrains the first derivative, ensuring it approaches zero along the \zaxis:
\begin{equation}
    \loss_{\temperatureEst_z} = \frac{1}{| \dataset_z |} \sum_{i=1}^{\dataset_z}  \temperatureEst_{z}^2 \qquad \text{and} \qquad \loss_{\temperatureEst_r} = \frac{1}{| \dataset_r |} \sum_{i=1}^{\dataset_r} \temperatureEst_{r}^2 \eqDot 
\end{equation}
The complete loss is defined as:
\begin{equation}
    \loss = \paramPDE (\lossPDE^\temperature +\lossPDE^\displacement) + \paramIC (\lossIC^\temperature + \lossIC^\displacement) + \paramBC (\lossBC^\temperature + \lossBC^\displacement) + \lambda_\posTS \loss_\posTS + \lambda_{\nuggetDia} \loss_{\nuggetDia} + \loss_{\temperatureEst_z} + \loss_{\temperatureEst_r} \eqDot
    \label{eq:complete_loss_2d}
\end{equation}
Here, the hyperparameters $\lambda$ weight the individual loss terms. Weighting the contribution of each individual loss term is achieved by weighting, which is crucial when terms vary in magnitude or importance.
\textcite{Wang2021} did not define a $\paramPDE$, reasoning that the collocation point is the main objective, whereas \textcite{Cuomo2022} used hyperparameters on all terms of $\loss$. \fig~\ref{fig:goal_losses} illustrates the losses.
The loss function $\loss$ serves as a means to enforce the adherence of the \acrlong{NN} to the \acrshort{PDE}, \acrlong{BC}, and observed data. Subsequently, the optimization criterion for \equ~\ref{eq:complete_loss_2d} can be expressed with $\hat{\params}$ as the estimated parameters of the \acrlong{NN}: $ \hat{\params} = \argmin_{\params} \left\{ \loss \right\}$.
%

%
%
The dataset consists of domain, initial, boundary, and aggregated experimental data. The dataset is partitioned into training, validation and test sets.
The collocation points of $\datasetPDE$ consist of space and time domains and are expanded with the process parameters $\{ r, z, t, \currentMax, \forceMax \}$. We use \lhs\ \autocite{McKay1979} to sample more representative random uniform samples than independent, identically distributed sampling \autocite{Bergstra2012}. It has been utilized by \textcite{Raissi2020} to sample collocation points from each $\body$.
The \acrlong{IC} dataset is defined as predictor-response pairs $\datasetIC \vcentcolon= \{ \predictor_\text{IC}, \response_\text{IC} \}$, where $\predictor$ is the predictor and $\response$ the response.
The boundary conditions are evaluated on $\datasetBC \vcentcolon= \{ \predictor_\text{BC}, \response_\text{BC}  \}$ on the boundary $\electrodeSheetBC$ to enforce the \neumann\ conditions.

The experiments span 22 current and four force levels, yielding 88 combinations. For each combination, 30 repetitions are welded. Each experiment measures dynamic displacement and nugget diameter. From these repetitions, 88 mean values are calculated. We disregard the first weld spot due to electrical contact irregularities, as the electrode caps are redressed before each sheet.
The target dynamic displacement is denoted as $\response_{\posTS}$ and the nugget diameters as $\response_{\nuggetDia}$. The dataset is structured by organizing the predictors $\predictor$ and the responses $\response$ variables according to individual welding experiments.
Neither target has the same shape as $\datasetPDE$, which is defined by its space-time domains. The targets have the shape of time for $\response_{\posTS}$, and a single time step at the end of the welding time $\zeit=560$ for $\response_{\nuggetDia}$. During training, the displacement field can be aggregated to match the dimension of $\response_{\posTS}$, and the temperature field can be evaluated on the specific points of $\response_{\nuggetDia}$. An example of both targets is:
\begin{align}
        \response_\posTS = \begin{bmatrix}
        \displacement(z=1, t=0) \\
        \displacement(z=1, t=1) \\
        \ldots \\
        \displacement(z=1, t=760) \\
        \end{bmatrix} \eqCom \quad \response_{\nuggetDia} = \begin{bmatrix}
        \temperature(r=0.0, z=0.5, t=560) \\
        \temperature(r=0.1, z=0.5, t=560) \\
        \ldots \\
        \temperature(r=1.0, z=0.5, t=560) \\
        \end{bmatrix}
        \eqDot
    \label{eq:dataset_matrix}
\end{align}
Concerning the evaluation protocol, the dataset is divided into 60\perc\ training, 20\perc\ validation, and 20\perc\ test sets based on individual welding experiments. This procedure ensures that data points from the same weld are not split across different welds, thereby preventing information leakage between welds, as the data is fully contained within each weld. The training set $\datasetTrain$ is used to fit the model, the validation set $\datasetVal$ is used for hyperparameter optimization, and the test set $\datasetTest$ is reserved for evaluating the final model. This method ensures the evaluation reflects the ability to generalize to unseen welds, which is an advantage compared to numerical simulation.
An important consideration for \acrshort{PINN}s is the setup of the datasets. For example, the initial condition $\datasetIC$ may be incorporated within the \acrlong{BC} with $\datasetIC \in \datasetBC$ \autocite{Raissi2017a}. We find that explicit datasets for the initial, boundary, and domain points enable a more transparent interpretation of the loss terms and model convergence.

\subsection{Training strategies for integrating experimental data}
This section introduces a method for dynamically updating temperature-dependent material parameters using the network's predicted temperature. We further discuss a delayed loss activation to mitigate training instability during early training. Subsequently, we present a progressive data assimilation approach that gradually introduces experimental goal losses. Lastly, we propose a rolling window for the learning rate scheduler and early stopping, which prevents premature convergence issues.

\subsubsection{Temperature-dependent material parameter update and delayed loss activation}
\label{sec:material_param_update}
When temperature-dependent material parameters are available, they must not be inferred by the PINN. 
However, the material parameters are not always available as a continuous analytical function that is easily differentiated.
An alternative is to provide the parameters as a lookup table. To incorporate the experimental values with a lookup table, the product and chain rule must be applied to separate the terms. For example, in the axial term of \equ~\ref{eq:heat_eq_loss}, the chain rule is expanded to:
\begin{equation}
    \pdv{}{z} \left( \thermCond(\temperature(z)) \pdv{ \temperature}{ z} \right)
    = \pdv{\thermCond(\temperature(z))}{\temperature} \left( \pdv{\temperature}{z}\right)^2 + \thermCond(\temperature(z))\pdv[2]{\temperature}{z} \eqDot
\end{equation}
Automatic differentiation can solve the \acrlong{LHS} of this equation, but requires $\lambda(T)$ to be a continuous function, which is not available. We use the \acrlong{RHS} with the defined functions using \equ~\ref{eq:material_parameter} and the derivatives with respect to $\temperature$ using look-up tables.
The material parameters can be updated using $\temperatureEst$ by matching the current values to their distinct room temperature values. Our proposed procedure dynamically updates the material parameters for each instance in a batch, $i \in \batch$, during each training step. This corresponds to a specific point $(r, z, \zeit)$, where the \acrshort{PINN} estimates the temperature $\temperatureEst$. For example, $\electrThermDissipation(\temperature)$ is included in the model and updated to $\electrThermDissipationEst$ given $\temperatureEst$ for each collocation point $i$. We define the update procedure as:
\begin{equation}
    \hat{\lambda}_\text{th}^i = \thermCond \left( \temperatureEst^i \right) \eqCom \qquad
    \electrThermDissipationEst^i = \electrThermDissipation \left( \temperatureEst^i \right) \eqCom \qquad
    \thermExpEst^{i} = \thermExp \left( \temperatureEst^i \right) \eqDot
    \label{eq:update_diffusivities}
\end{equation}
%

%
%
During the initial stages of training, the network predicts unrealistic temperatures and displacements. The network predicts all possible values, including negative temperatures, which distorts the optimization process. As material parameters are temperature-dependent, unreasonable values would be used at the very beginning of the training. For example, negative temperatures must be handled by clamping or extending the material parameter values. 
In our implementation, the network is initially trained to predict reasonable values for $\temperatureEst$, after which the material properties are updated. This approach helps avoid the instability caused by unreasonable temperature estimates early in training.
We delay the activation of \equ~\ref{eq:update_diffusivities}. This strategy is inspired by the approach of \textcite{Zhang2022PINN} and \textcite{Moseley2020}, where $\lossPDE$ is delayed.
Rather than using a fixed training step for activation, we employ a threshold value for the estimated temperature, $\lossPDEThreshold$, also illustrated in \fig~\ref{fig:PINN_RSW_material_model}. The material parameters are added to the optimization once the following condition is fulfilled:
\begin{equation}
    \loss \le \lossPDEThreshold \eqDot
\end{equation}

In addition, a loss function is introduced to prevent the \acrshort{PINN} from predicting negative temperatures in general. During model development, it was observed that $\temperatureEst$ was occasionally below zero during the initial stages of training. This issue arises in the context of competing objectives within $\lossPDE$. To safeguard the prediction, the \acrshort{PINN} is constrained by imposing a penalty for $\temperatureEst \lt 0$. The loss is added to the total loss in \equ~\ref{eq:complete_loss_2d}, without a hyperparameter. The points are sampled from $\datasetPDE$ and preserve the \acrlong{AD} by masking the prediction array against negative infinity:
%
\begin{equation}
    \lossTemp = \frac{1}{| \datasetPDE |} \sum_{i=1}^{| \datasetPDE |} \left[ \min \left\{ 0, \max \left\{ \temperatureEst_i, -\infty \right\} \right\} \right]^2 \eqDot
    \label{eq:negative_loss}
\end{equation}
\subsubsection{Progressive experimental data assimilation}
Instead of introducing the loss terms for the experimental data abruptly, we employ a scaling factor $f$ that gradually fades in the additional loss following a sigmoid-shaped function. The gradual introduction of additional loss terms ensures a smooth transition, resulting in minimized contribution from the added loss at the early stage. This enables the network to adapt without destabilizing gradients or hindering convergence because the network does not have to handle competing objectives. An example of an abrupt introduction is given \fig~\ref{fig:proof_smoothing_and_rolling_window}.
The experimental data losses, \ie\ $\{ \loss_{\nuggetDia}, \loss_{\posTS}\}$, are added to the optimization when the threshold is undercut, as previously argued. The fading-in process begins when the threshold $\temperatureThreshold$ is undercut, denoted as the reference epoch $\referenceEpoch$. The fading duration is defined with a number of transition epochs $\transitionEpoch$ during which the partial loss is denoted as $\lossGoal^\prime$. Here, the scaling factor is defined with the steepness of the sigmoid curve set to 0.01:
\begin{equation}
    \lossGoal^\prime = f \times \lossGoal \qquad \text{where} \qquad f = \frac{1}{1 + e^{-0.01 (\epoch - \referenceEpoch - \transitionEpoch)}} \eqDot \label{eq:smoothing_factor}
\end{equation}
\subsubsection{Rolling window learning rate scheduler and early stopping}
Common types of learning rate schedulers track the performance of a metric and reduce the learning rate after a specified number of epochs with no improvement, or after a fixed step size, such as every 1,000 epochs. In performance-based approaches, a parameter known as patience is often employed to prevent overly frequent reductions in the learning rate. This is the number of epochs, $\numberOfBadEpochs$, during which the metric is allowed not to improve before triggering a reduction. This results in an issue when integrating experimental data because an additional goal loss increases the total loss. Ordinary learning rate schedulers often result in a rapid reduction of the learning rate, which can lead to training stalling. This refers to the situation where the learning progress severely decreases or stops completely.
We define a dynamic approach that leverages a rolling window with a length $k$ to mitigate this effect. This enables considering only losses within a rolling window from $\epoch - k$ to $\epoch$. The best loss during this rolling window is denoted as $\loss_\text{best}$:
\begin{equation}
    \loss_\text{best} = \min \{ \loss_{s-k}, \ldots, \loss_{s} \} \eqDot
\end{equation}
The new number of bad epochs $\numberOfBadEpochs^\prime$ is updated based on the most recent loss metric $\loss_s$ as follows:
\begin{equation}
    \numberOfBadEpochs^\prime = \left\{ 
        \begin{array}{cl}
            \numberOfBadEpochs + 1 & : \ \loss_s > \loss_\text{best} \\
            0 & : \ \loss_s \leq \loss_\text{best} \quad \text{or} \quad \numberOfBadEpochs = k
        \end{array} 
    \right.
    \label{eq:rolling_window}
\end{equation}
The learning rate is reduced only when $\numberOfBadEpochs$ exceeds the patience:
\begin{equation}
    \learningRate = \left\{
        \begin{array}{cl}
            \learningRate & : \ \numberOfBadEpochs < k \\
            0.1 \times \learningRate & : \ \numberOfBadEpochs = k
        \end{array} 
    \right.
\end{equation}
Here, an order of magnitude is chosen for reducing $\learningRate$. The main advantage is that the optimizer can retain a high learning rate while optimizing for additional optimization criteria, such as experimental data. Furthermore, stalling is prevented, and the functionality of reducing $\learningRate$ is maintained if the loss does not converge.
While the rolling window strategy offers advantages, it introduces additional computational overhead due to updating the set $\min \{ \loss_{\epoch - k}, \ldots, \loss_s \}$. However, this is worth the cost because the total training time is reduced.

The equivalent logic also applies to early stopping \autocite{Prechelt2012}, a regularization technique to prevent overfitting. In its original sense, early stopping detects plateaus or stalls in the validation loss and terminates training after a specified grace period, known as patience. The validation step in PINN training would sample domain or boundary points from the same domain as the training step. Additionally, frameworks often disable gradient backpropagation during validation because the network's weights are not updated, rendering training of PINNs highly inefficient when backpropagation is deliberately allowed. We monitor the PINN loss to detect a plateau and stop the training before the model diverges. The early stopping must also be defined using a rolling window. The increased loss from the newly added experimental losses might be interpreted as a lack of improvement, potentially triggering early stopping prematurely.

\section{Results and Discussion}
In this section, we begin by establishing a quantitative benchmark through training a PINN on a 1D heat equation with homogeneous \dirichlet\ boundaries and a sinusoidal heat source, for which an analytical solution exists. A benchmark loss value is established for the analytical solution, serving as a reference for assessing the PINN accuracy. Building upon this reference, we introduce a temperature-dependent material update procedure that utilizes a loss threshold. This section emphasizes the significance of controlled activation of material parameter updates to ensure stable and efficient training. Next, we explore hyperparameter optimization and training strategies using a 1D model. The initial focus on a one-dimensional approach is motivated by significantly lower computational costs without the overhead of large collocation grids. We demonstrate the effectiveness of the training strategies when incorporating additional experimental data constraints. We also compare inverse and forward PINN approaches for modeling contact heat generation and examine the resulting temperature fields and dynamic displacement in one dimension. After demonstrating these strategies in this controlled setting, we extend the model to a two-dimensional case, including the growth of the nugget diameters. By comparing it with the 2D variant, we discuss whether the 1D analysis is sufficient to describe aluminum RSW and thus the quality of the welding. All models used in this chapter are listed in \tab~\ref{tab:model_overview}, describing the achieved loss and their neural network parameters.
\begin{table}[h]
\centering
\begin{tabular}{llllll}
    \toprule
    & \textbf{Note} & \textbf{Section} & \textbf{Referenced Figures} &  \textbf{Train Loss} & \textbf{Params} \\
    \midrule
    \#1 & Analytical model & 4.1 & - & $3.01 \times 10^{-7}$ & $50 \times 2$ \\
    \#2 & Immediate $\loss_\posTS$ & 4.2.1 & \fig~\ref{fig:proof_smoothing_and_rolling_window} & $3.51 \times 10^{0}$ & $66 \times 3$ \\
    \#3 & No smoothing & 4.2.1 & \fig~\ref{fig:proof_smoothing_and_rolling_window} & $0.11 \times 10^{0}$ & $66 \times 3$ \\
    \#4 & No rolling window & 4.2.1 & \fig~\ref{fig:proof_smoothing_and_rolling_window} & $1.95 \times 10^{-2}$ & $66 \times 3$ \\
    \#5 & Smoothing and rolling & 4.2.1 & \fig~\ref{fig:proof_smoothing_and_rolling_window} & $6.53 \times 10^{-7}$ & $66 \times 3$ \\
    \#6 & Forward model & 4.2.2 & \fig~\ref{fig:paper_1_forward_vs_inverse} & $17.8 \times 10^{-6}$ & $66 \times 3$ \\
    \#7 & Inverse model & 4.2.2 &  \fig~\ref{fig:paper_1_forward_vs_inverse} & $1.30 \times 10^{-6}$ & $66 \times 3$ \\
    \#8 & 1D model & 4.2.3 & Figs. \ref{fig:paper_1_losses_1d_model},\ref{fig:paper_1_contact_param},\ref{fig:visualize_full_model_low},\ref{fig:visualize_full_model_high} & $1.30 \times 10^{-6}$ & $66 \times 3$ \\
    \#9 & 2D model & 4.3 & Figs. \ref{fig:paper_1_losses_2d_model},\ref{fig:paper_1_nugget_growth},\ref{fig:paper_1_2d_model_all_test},\ref{fig:paper_1_nugget_loss},\ref{fig:pinn_2d_pred_1},\ref{fig:pinn_2d_pred_2},\ref{fig:pinn_2d_pred_3}  & $6.02 \times 10^{-6}$ & $66 \times 3$ \\
    \bottomrule
\end{tabular}
\caption{Overview of trained models. Parameters describe the neural network layer size and number of layers.}
\label{tab:model_overview}
\end{table}

\subsection{Reference solution with \dirichlet\ boundaries and sinusoidal heat source}
%
The magnitude of the loss can serve as an indicator of the quality of the solution. To meaningfully interpret whether a loss value is sufficient, a reference loss is needed. Without a direct analytical or numerical reference, we use a simple \acrshort{1D} problem with a known analytical solution as a benchmark for the magnitude of the loss. The heat equation can be simplified to a one-dimensional, non-homogeneous partial differential equation subject to homogeneous Dirichlet boundary conditions. Specifically, the temperature at the boundaries is fixed, with $ \temperature(0, \zeit) = \temperature(1, \zeit) = 0$ while the initial temperature distribution is given by $\temperature(z, 0) = 0$. The resulting formulation of the problem is \autocite{Farlow1993}:
\begin{equation}
    \pdv{\temperature(z, \zeit)}{\zeit} = \thermDiffusivity \pdv[2]{\temperature(z, \zeit)}{z} + \sin ( \pi z ) \eqDot \label{eq:onedim_heat_eq}
\end{equation}
In this case, $\heatSupply$ is assumed to be constant within the domain $z \in [0, 1]$. We specify the heat source as $\sin(\pi z)$ to derive the analytical solution:
\begin{equation}
    \temperature(z, \zeit) = e^{-(\pi \thermDiffusivity)^2 \zeit} \sin(\pi z) + \frac{1}{(\pi \thermDiffusivity)^2} \left( 1 - e^{-(\pi \thermDiffusivity)^2 \zeit} \right) \sin( \pi z) \eqDot
    \label{eq:simple_analytical_solution}
\end{equation}
We train a \acrshort{PINN} to solve the \dirichlet\ setup to obtain a benchmark loss. The one-dimensional problem is defined as:
\begin{align}
    \lossPDE &= \frac{1}{| \datasetPDE |} \sum_{i=1}^{| \datasetPDE |} \left[ \pdv{\temperatureEst_i}{\zeit} - \thermCond \pdv[2]{\temperatureEst_i}{z} - \sin(\pi z) \right]^2 \eqCom \\
    \lossIC &= \frac{1}{| \datasetIC |} \sum_{i=1}^{| \datasetIC |} \temperatureEst_i^2 \eqCom \quad \lossBC =  \frac{1}{| \datasetBC |} \sum_{i=1}^{| \datasetBC |} \temperatureEst_i^2 \eqDot
\end{align}
The model achieves a reference loss of $3.01 \times 10^{-7}$, at which no visual difference to the analytical solution is observed.

\subsection{Test of training strategies for integrating experimental data in one dimension}
%
%
%
Since it is unknown which set of hyperparameters is optimal for our learning task, exploring a wide range of hyperparameters is usually necessary. With a number of weighting terms in \equ~\ref{eq:complete_loss_2d}, this exploration leads to a hyperparameter optimization problem. This optimization is a formal process \autocite{Bergstra2011} of searching the hyperparameter space, in which we make as few prior assumptions as possible, thereby leaving the selection to the algorithm. 
\textcite{Bergstra2012} experimentally showed that exploring the search space randomly is a favorable approach compared to manual or grid search.
All continuous hyperparameters are defined on a logarithmic scale to span several orders of magnitude efficiently. We use \hyperopt\ \autocite{Bergstra2013} to suggest configurations according to a random sampling strategy, outperforming manual and grid searches in practice \autocite{Bergstra2012}.
The neural network size is explored, ranging from two to six hidden layers, each with 8 to 512 neurons. The weighting of the terms of each loss term is searched on a scale between $10^{-2}$ and $10^1$.
We focus on three key hyperparameters for the network: the optimizer, which governs how the network’s weights are updated during training; the activation function, which introduces non-linearity to the model; and the weight initialization method, which sets the initial values of the weights.
To evaluate how different optimization dynamics affect convergence, we investigate four popular algorithms: \adam, \adadelta, \acrshort{RMSprop}, and \mbox{limited-memory} \acrfull{BFGS}.
The choice of the activation function is an active field of research with no conclusive consensus. Recognizing that activation functions are crucial to model stability and training, we examine \tanH, \acrshort{GELU}, and \acrshort{ELU}. \acrshort{ReLU} and \leakyReLU\ are not included because, as \textcite{Sitzmann2020} noted, it breaks the second derivative. 
Since the initial distribution of network weights can influence gradient behavior at the start of training, we compare two widely used initialization schemes: \xavier\ initialization and \kaiming\ initialization, using both normal and uniform distributions.
Furthermore, we do not use batch normalization as it has been shown to break the meaning of the derivative. The search space is summarized in \tab~\ref{tab:hyperparameter}.
The hyperparameter optimization for the \acrshort{1D} setup used $\datasetTrain$ and was evaluated on $\datasetVal$. This resulted in a model with three hidden layers and 66 neurons. In fact, increasing the number of neurons consistently decreased the final loss, suggesting potential overfitting. Although previous studies \autocite{Raissi2017a} have utilized the \acrshort{BFGS} optimizer, the hyperparameter results suggest that the Adam optimizer offers smoother convergence and facilitates the implementation of training strategies. 
The layer initialization has little effect, with \xavier\ having an edge. The $\tanh$ activation function, a threshold of $\lossPDEThreshold = 0.001$, and an initial learning rate of $\learningRate = 10^{-3}$ were determined to be the most effective for our application. The number of data points is 10,000, and the regularization parameters were set to $\lambda_\posTS = 0.7$ with all other parameters set to one.
\begin{table}
\centering
\begin{tabular}{lccc}
    \toprule
    \textbf{Hyperparameter} & \textbf{Value range} \\
    \midrule
    number of hidden layers             & $[2; 6]$ \\
    size of hidden layers                   & $[8; 512]$ \\
    optimizer                   & \makecell{\adam, \adadelta, \acrshort{RMSprop}, limited-memory \acrshort{BFGS}} \\
    layer initialization        & \makecell{(normal or uniform) \xavier\ or \kaiming} \\
    activation function         & \makecell{\tanH, \acrshort{GELU}, \acrshort{ELU}} \\
    initial learning rate $\learningRate$           & $[10^{-6}; 10^{-1}]$ \\
    number of datapoints $| \datasetPDE |$  & $[10^3; 10^5]$ \\
    $\lambda_\text{PDE}$, $\lambda_\text{IC}$, $\lambda_\text{BC}$, $\lambda_{\posTS}$, 
 $\lambda_{\nuggetDia}$ & $[10^{-2}; 10^1]$ \\
    \bottomrule
\end{tabular}
\caption{Hyperparameter search space.}
\label{tab:hyperparameter}
\end{table}
\subsubsection{Smoothing factor and rolling windows}
\begin{figure}
    \centering
    \includegraphics[width=\imgWidth\linewidth]{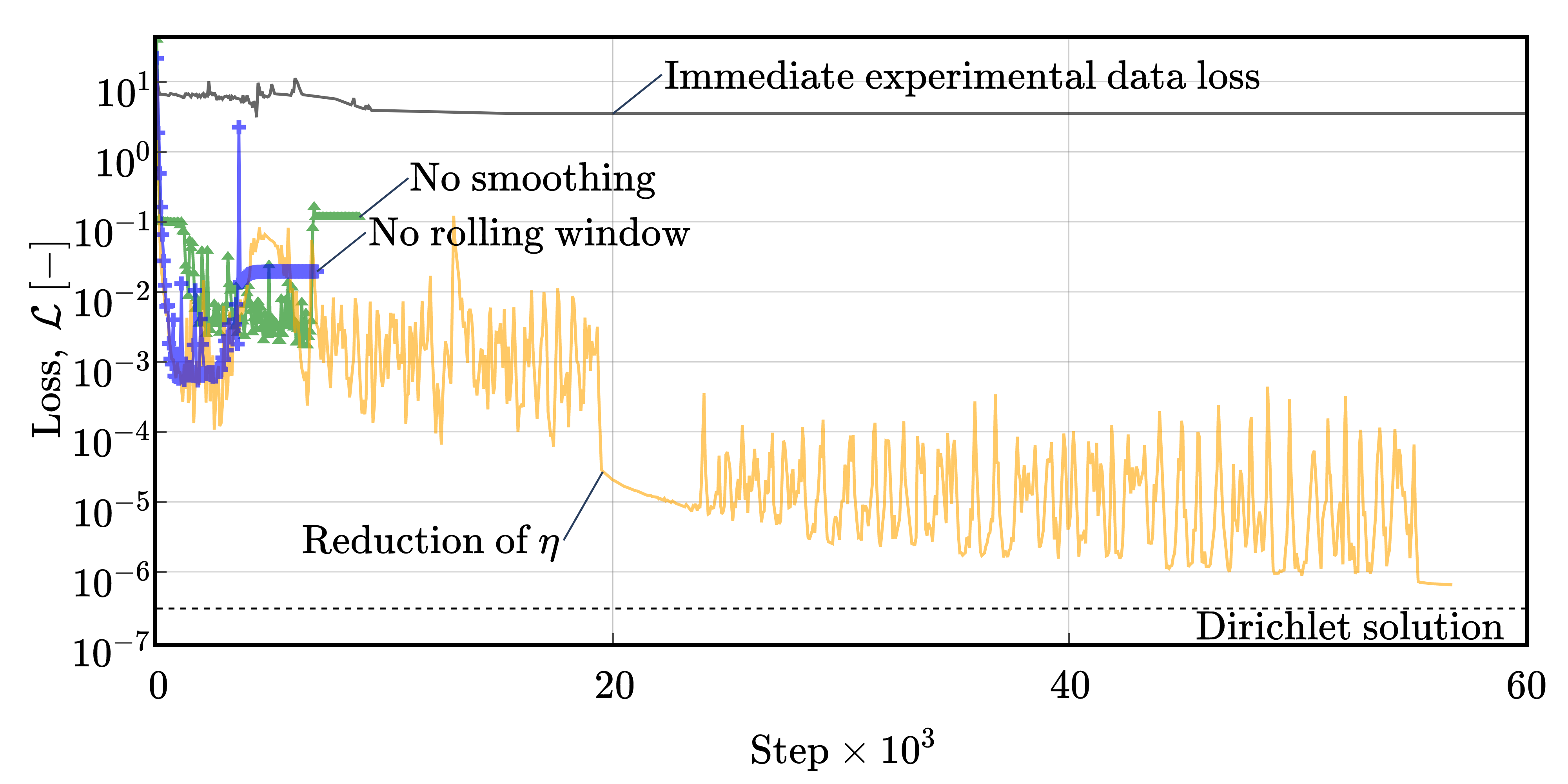}
    \caption{Effect of smoothing factor and rolling window on separate models on mockup goal losses. The exact reference epoch depends on when the threshold $\lossPDEThreshold$ is undercut. Grey: Immediate activation of the experimental goal loss, resulting in no convergence and eventually in stalling. Green: The abrupt introduction of additional optimization criteria can result in stalling. Blue: Gradual introduction results in stalling because the learning rate is reduced too fast. The training is stopped. Yellow: A rolling window allows for a relatively high learning rate while balancing the optimization criteria.}
    \label{fig:proof_smoothing_and_rolling_window}
\end{figure}
As previously discussed, the abrupt introduction of additional constraints, such as time series loss, $\loss_\posTS$, frequently leads to optimization conflicts and stalling in PINN training. We investigate the effect of two key strategies: smoothing the introduction of the new loss terms and employing a rolling window for the learning rate scheduler and early stopping. Both strategies are defined in \equ~\ref{eq:smoothing_factor} and \equ~\ref{eq:rolling_window}. This part focuses on integrating a goal loss, which is why the models are trained on a mockup goal loss with a heat source defined in \equ~\ref{eq:contact_res_heat}. The mock-up goal loss is set twice as high as the default solution without goal loss. The loss is later replaced by the experimental dynamic displacement and is referred to as $\loss_\posTS$.
We introduce three distinct models to evaluate the effects. The first model, which immediately activates $\loss_\posTS$, without waiting for a $\lossPDEThreshold$, illustrates the need for the threshold.
The second model, which uses no smoothing, highlights the immediate spike in loss that results from the abrupt addition of a constraint.
The third uses smoothing but no rolling window.
The fourth model combines both strategies, where we initially trained the \acrshort{PINN} to solve the governing equation and boundary conditions. Once $\lossPDEThreshold$ is met, the time series loss $\loss_\posTS$ and material parameter update are added to the optimization. As argued before, when the additional constraints are added abruptly, the training almost always stalls immediately. This is because an optimization conflict arises between the introduced data objective and the existing PDE and boundary‐condition objectives. This conflict is visible with the green trace in \fig~\ref{fig:proof_smoothing_and_rolling_window}, where the training stalls after adding $\loss_\posTS$. To overcome this, we gradually introduced the constraints using the smoothing function defined in \equ~\ref{eq:smoothing_factor} over 1,500 epochs. This allows a smooth, continued convergence of the other losses, \eg\ $\lossPDE$ and $\lossBC$. The blue trace shows the effect of the smoothing function.

A secondary issue is that the total $\loss$ inevitably remains above the lowest previous value for an extended time. This prematurely triggers the learning rate scheduler and early stopping. We addressed this by implementing a rolling window, \cf\ \equ~\ref{eq:rolling_window}, with a window size equal to the scheduler's patience parameter, which is optimal. In our models, the number of steps is 1,000. The strategy enables the training to begin with comparatively high learning rates, at $\learningRate = 10^{-3}$, while maintaining loss convergence.
The rolling window only considers the last $k + 500 = 1,500$ steps. Training is terminated if no decrease in loss is observed within the entire window.
The yellow trace in \fig~\ref{fig:proof_smoothing_and_rolling_window} shows a gradual increase in loss, followed by a rapid return to the previous level of loss. The network’s loss steadily decreases to values much closer to the \dirichlet\ benchmark. These results indicate that combining a smoothing term and a rolling window yields improved convergence behavior and lower final loss than abrupt constraint introduction.

\subsubsection{Comparison of an inverse and forward PINN for contact heat generation}
%
%
%
\begin{figure}
    \centering
    \includegraphics[width=\imgWidth\linewidth]{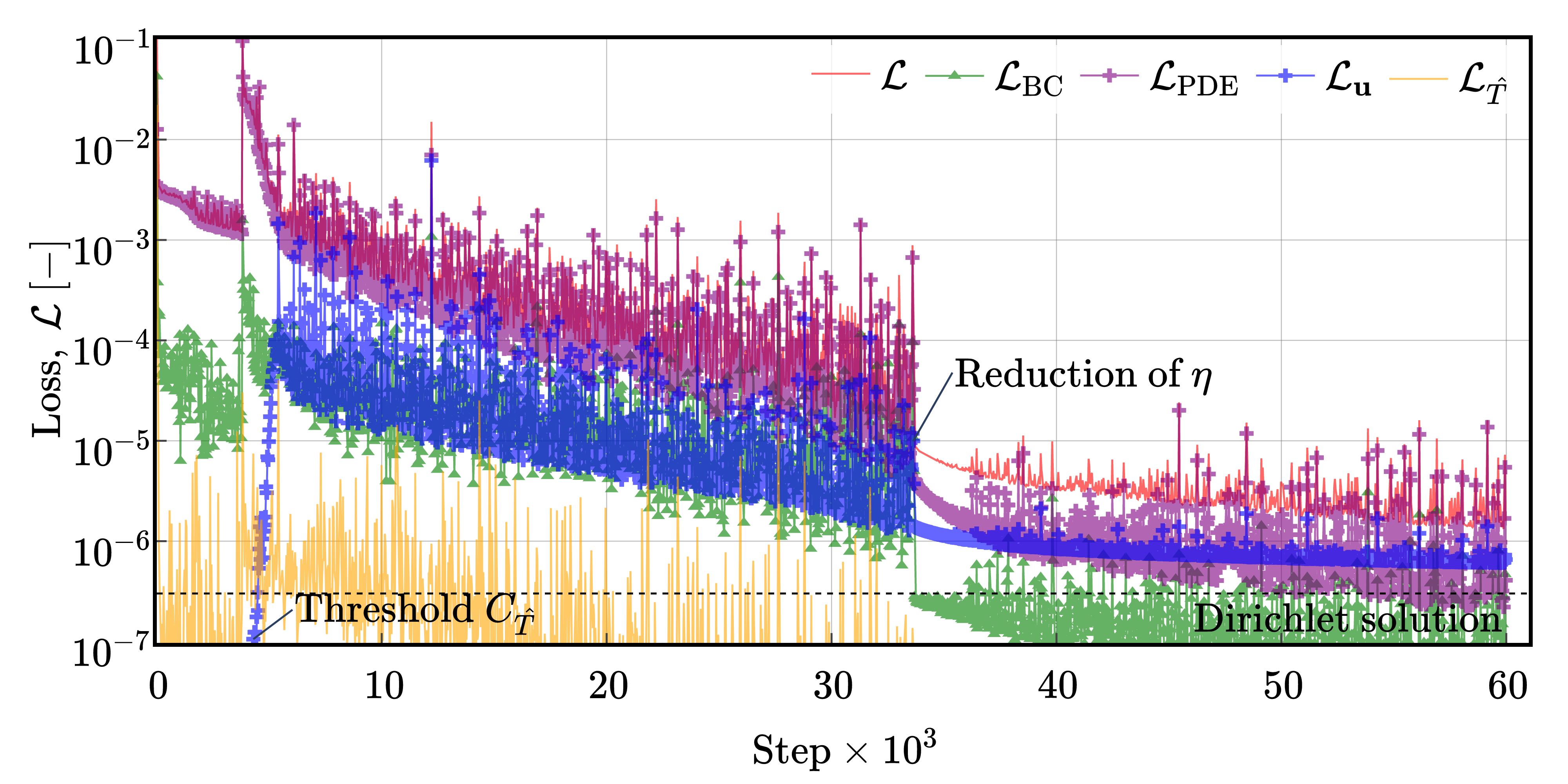}
    \caption{
    Loss convergence of the 1D model. The plot illustrates the convergence of the total loss and its contributing components: boundary condition, PDE, experimental displacement, and penalty on negative losses. The Dirichlet solution is presented for reference. 
    The fade-in of the displacement loss is visible around 4k epochs. The overall convergence follows a linear trend until \~32k epochs, where the first learning rate reduction occurs. A final learning rate reduction at the end is barely visible.}
    \label{fig:paper_1_losses_1d_model}
\end{figure}
\begin{figure}
    \centering
    \includegraphics[width=\imgWidth\linewidth]{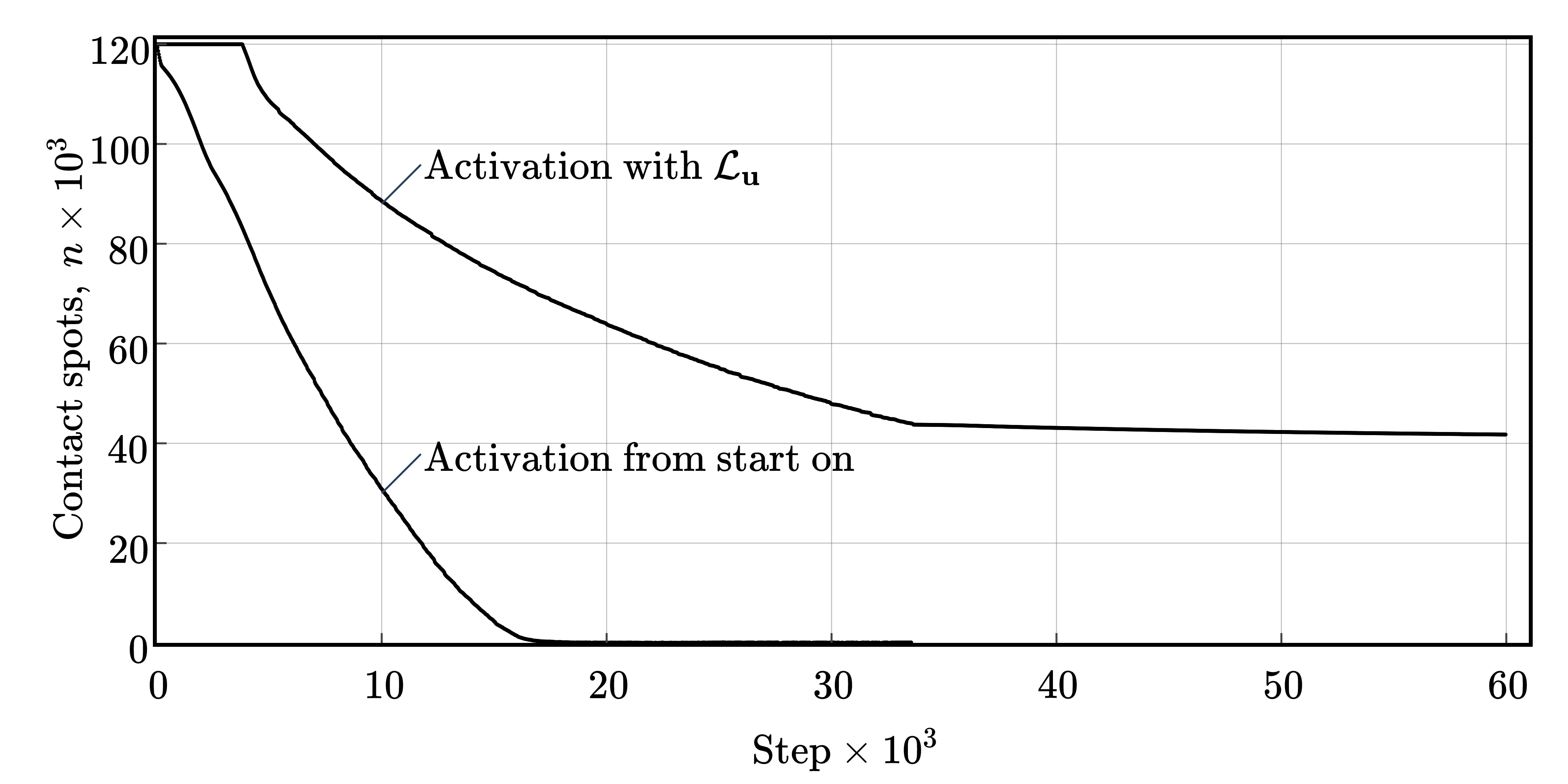}
    \caption{
    Optimization strategies for the network parameter $n$. Early activation, from the start on, to the optimizer results in $n \approx 90$, leading to inflated constriction resistance. The value remains about the $n<1$ threshold for hyperbolic growth. The convergence is not affected once the experimental loss term is added. Delayed addition, alongside experimental data loss, leads to $n \approx 40,000$.
    }
    \label{fig:paper_1_contact_param}
\end{figure}
We train a complete 1D model on $\datasetTrain$ and validate using the prediction on $\datasetTest$ and the confidence interval of the experimental data. The model learns from the real-world experimental dynamic displacement data. Training is allowed up to $10^5$ epochs, but early stopping ends the training if no improvement is detected or the analytical reference loss is achieved: $\loss <= 3.01 \times 10^{-7}$. Loss convergence was achieved despite a spike once $\lossPDEThreshold$ was undercut, as shown in \fig~\ref{fig:paper_1_losses_1d_model}. In other experiments without the smoothing strategy, the spike was observed to be higher by a factor of $10^3$, resulting in immediate stalling, as evident in \fig~\ref{fig:proof_smoothing_and_rolling_window}. This spike is crucial to the model's convergence. The training strategies effectively balance the spike and yield better results, as indicated by the convergence of $\lossPDE$. Other loss spikes around $\loss = 10^{-4}$ are not uncommon and align with findings from various studies, where sharp increases in loss are typically followed by rapid recovery. This behavior has been observed across a range of model architectures and applications, as noted in the works of \textcite{He2023,Hennigh2021,Zhang2022PINN}.
In addition, $\loss_{\temperatureEst}$ vanishes after 34,000 epochs, and the training terminates after the last learning rate reduction. Those kinds of spikes are not attributed to a specific type of loss. The training does not achieve \dirichlet\ accuracy but produces continuous and reasonable temperature and displacement fields.
The model strikes a balance between the optimization criteria. The employed training strategies facilitate efficient convergence. It is essential to note that all separate losses converge, indicating an effective model.

Using \equ~\ref{eq:contact_res_heat} to model the contact heat, the optimization of the number of contact spots, $n$, is of particular interest because this parameter influences heat generation at the faying surface.
Initial experiments, where $n$ was a learnable parameter from the onset of training, resulted in models that aggressively minimized $n$. This behaviour led to $n$ converging to unreasonable low values around $n \approx 90$, which in turn inflates the constriction resistance part of $\contactRes$. Notably, this value is still above the threshold of $n \le 1$, below which the contact heat would become hyperbolic. Reconsidering the effect of defining contact resistance with $n$ contact spots, it underestimates constriction resistance by $\sqrt{n}$ compared to film resistance. We reason that this is the result of the optimizer prioritizing $n$, even as the experimental loss $\loss_{\posTS}$ is introduced. 
To achieve a more physically sound model, $n$ is added to the optimizer once $\loss_{\posTS}$ is introduced. This delay resulted in a convergence to a value of 40,000, effectively more than halving the initial estimate, but still well above 90. The resulting convergence is shown in \fig~\ref{fig:paper_1_contact_param}. 
We reason that $\loss_\posTS$ provides a desirable optimization conflict, which is mediated by $n$. 
Still, the initial estimate is off and convergence of $n$ suggests that either the number of initial contact spots $n$ was overestimated or other variables in \equ~\ref{eq:contact_res_heat}, \eg\ film thickness or film resistance, were assumed too low.
Considering that the variance in film resistance spans orders of magnitude \autocite{Tang2013,Mousavi2022}, and the film thickness is only assumed to be of the order of magnitude \autocite{Wang2001}, this seems reasonable. In addition, adhesive remains may influence the contact parameter, as argued by \textcite{Zhao2018}.
\begin{table}
\centering
    \begin{tabular}{lllllll}
    \toprule
    & \textbf{Train Loss} & \textbf{Test R\textsuperscript{2}}  \\
    \midrule
    \textbf{Forward PINN} & $17.8 \times 10^{-6}$ & 0.985 \\
    \textbf{Inverse PINN} & $1.3 \times 10^{-6}$ & 0.974 \\
    \bottomrule
    \end{tabular}
\caption{Train loss and accuracy of a forward and inverse \acrshort{PINN}.}
\label{tab:inverse_vs_forward}
\end{table}

\begin{figure}
    \centering
    \includegraphics[width=\imgWidth\linewidth]{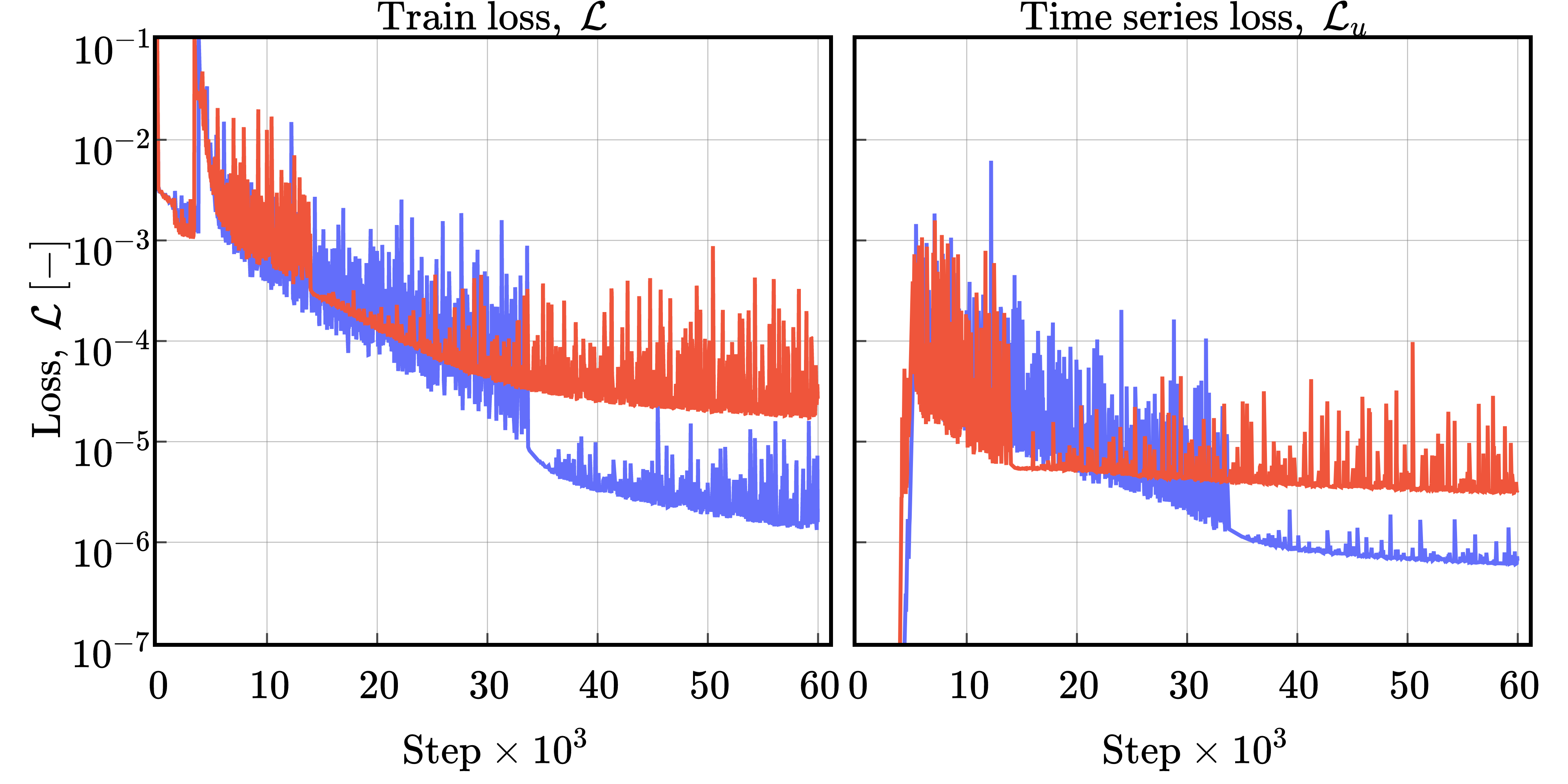}
    \caption{
    Comparison of a forward and inverse \acrshort{PINN}. The blue curve reflects the losses for an inverse and the red curve for a forward PINN. Both models reduce the learning rate once. The inverse model converges to a loss about 8 times lower.
    }
    \label{fig:paper_1_forward_vs_inverse}
\end{figure}

\begin{figure}
    \centering
    \includegraphics[width=\imgWidth\linewidth]{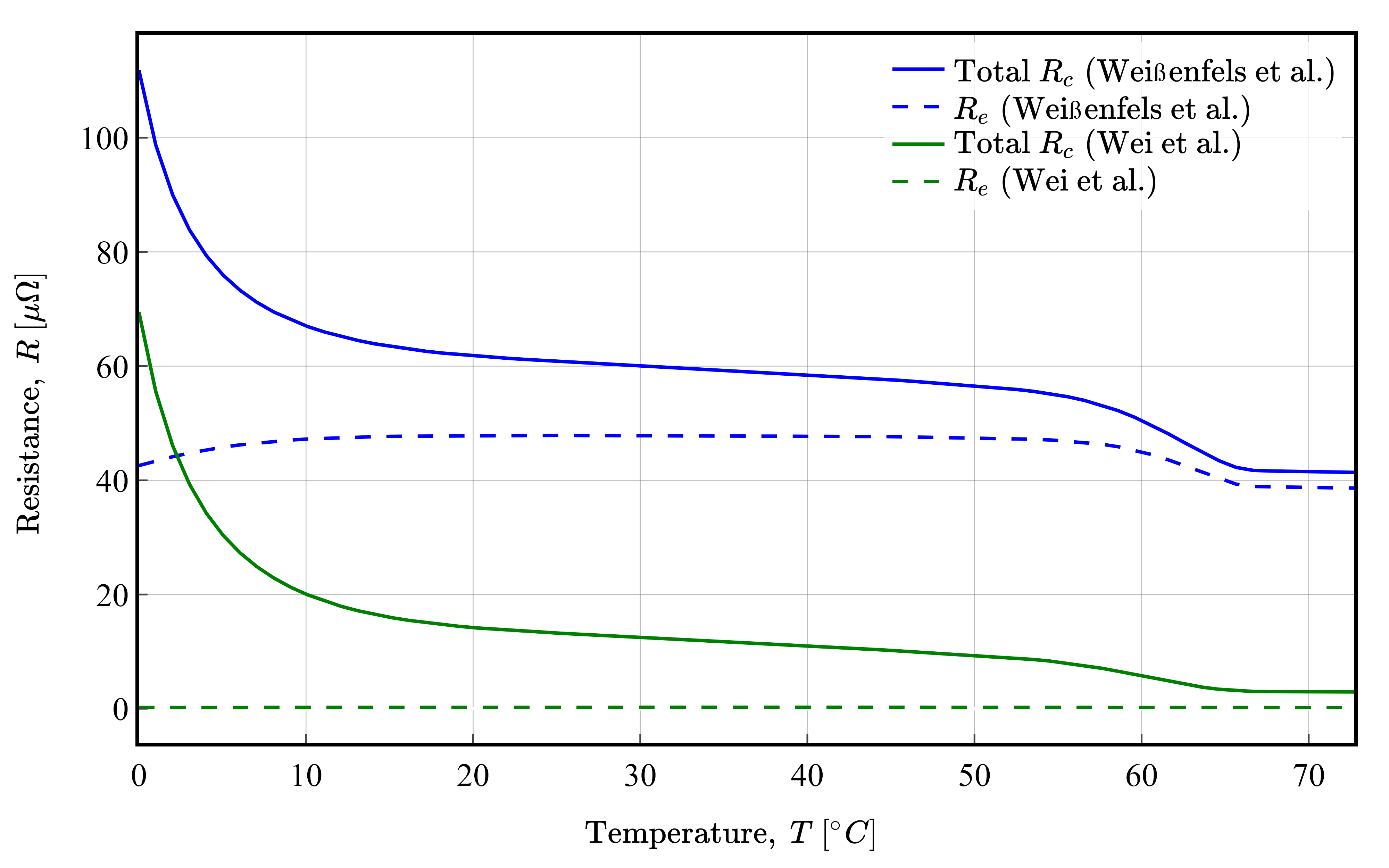}
    \caption{
    Total contact resistance and its constriction component for the contact equation used in the inverse and forward one-dimensional models. Constriction resistance is calculated from the estimated temperature at each time step. The forward model shows a significant constriction component, while the inverse model estimates a negligible, near-zero constriction resistance.
    }
    \label{fig:contact_resistance_film_const}
\end{figure}

Considering the contact parameter, the question arises whether forward training would save computational resources, reduce optimization complexity, and produce the same result.
We compare the inverse approach with a forward \acrshort{PINN} that uses \equ~\ref{eq:weissenfels} as contact resistance. This comparison can reveal which approach is better for modelling contact heat generation. We evaluate the two methods based on the loss convergence and test accuracy.
A forward model is trained using the contact heat generation defined in \equ~\ref{eq:contact_res_heat_B}. This formulation does not require the estimation of $n$, which can benefit model training. It relies on the constriction resistance and considers a spreading resistance factor.
Quantitatively, the resulting loss and accuracy are summarized in \tab~\ref{tab:inverse_vs_forward}. The $\rSquared$ has been calculated over 17 time series from $\datasetTest$. \fig~\ref{fig:paper_1_forward_vs_inverse} compares the loss convergence of both model choices. Across several experiments, it is consistently observed that $\loss_\posTS$ does not converge as fast or to the same magnitude as the inverse PINN.
We deduce that $n$ facilitates convergence by mediating the optimization conflict. However, the forward model is expected to achieve the same result, considering the dynamic displacement.
The contact resistance can be calculated backwards from the predicted temperature at the faying surface. \fig~\ref{fig:contact_resistance_film_const} shows the predictions of the inverse and forward models. It is evident that the contact resistances contain different proportions of constriction resistance. 
The inverse model results in a lower loss. This suggests that the equations were better solved by the inverse model, which effectively underestimated the constriction resistance.
%

%
\subsubsection{Temperature fields and dynamic displacements}
\begin{figure}[!hb]
    \centering
    \includegraphics[width=0.85\linewidth]{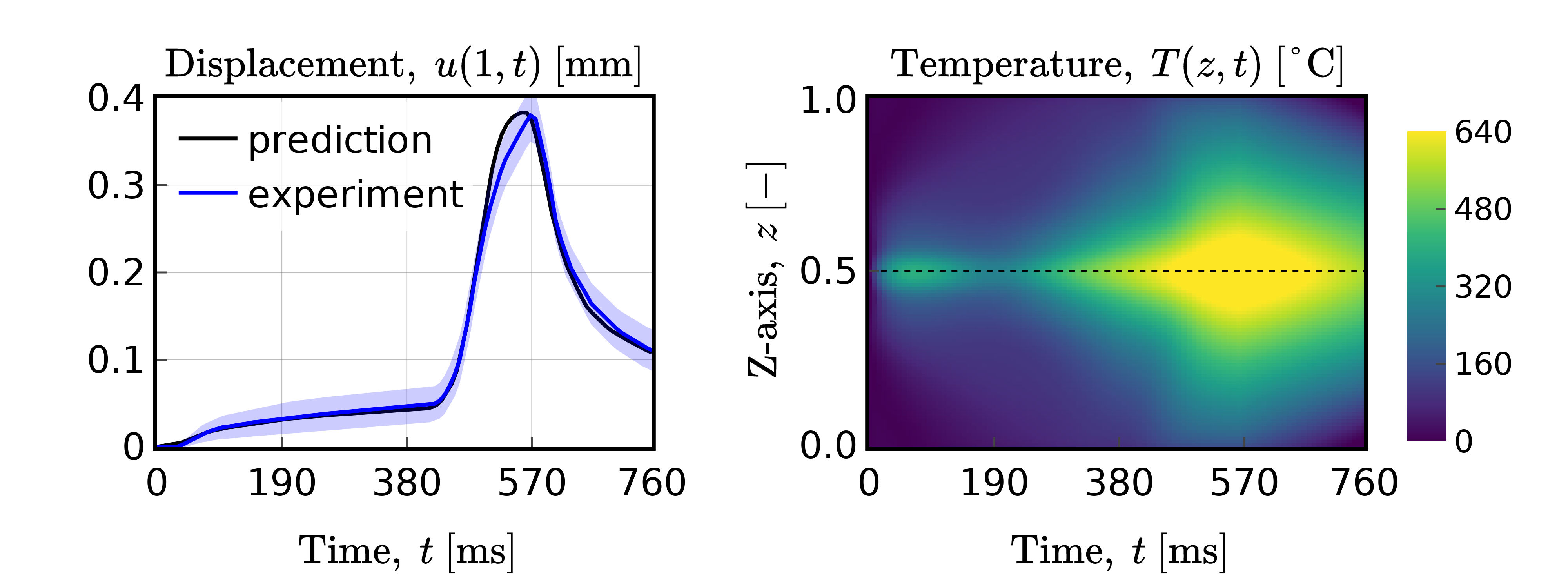}
    \caption{Predicted dynamic displacement and temperature field of the 1D model on unseen test data.}
    \label{fig:visualize_full_model_low}
\end{figure}
\begin{figure}[!hb]
    \centering
    \includegraphics[width=0.85\linewidth]{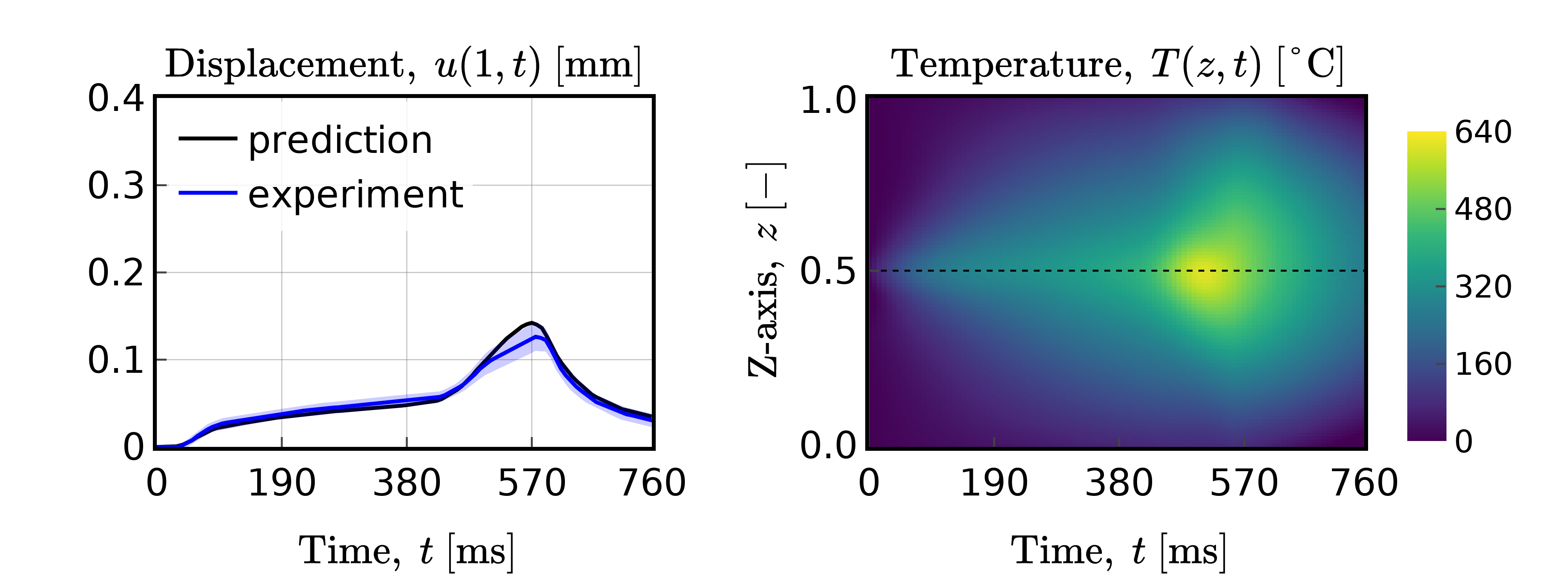}
    \caption{Predicted dynamic displacement and temperature field of the 1D model on unseen test data.}
    \label{fig:visualize_full_model_high}
\end{figure}
%
Simulating different current and force levels shows the model's predictive capabilities. Two extremes are chosen from $\datasetTest$ to illustrate the range of the temperature fields. The upper extrema is a high $\currentMax = 47$ with a low $\forceMax = 5$, and the lower extrema is $\currentMax = 26$ with $\forceMax = 8$. Those yield the highest and lowest $\posTS$ and $\temperature(z, \zeit)$ for the \acrlong{AL6} combination. The temperature fields are shown in \fig~\ref{fig:visualize_full_model_low} and \fig~\ref{fig:visualize_full_model_high}. The temperature evolution during preheat and hold time is also similar, especially when examining the edges. It is observed that the model handles discontinuity from the welding schedule at $\zeit = 560$ well, \cf\ \equ~\ref{eq:stromprofil}. 
Regarding the displacement field, $\posTS$ is compared to the experimental variance of the data. Recall that one sheet of 30 spots was welded with the same parameters, giving a mean and standard deviation. The mean was used for training the network, while the standard deviation was used to assess its performance. The predicted $\posTS$ lies within the confidence interval with an overall accuracy on $\datasetTest$ of $\rSquared = 0.97$. The prediction was shown to be smoother than the experimental mean.

\subsection{Two-dimensional aluminum resistance spot welding}
\begin{figure}
    \centering
    \includegraphics[width=\imgWidth\linewidth]{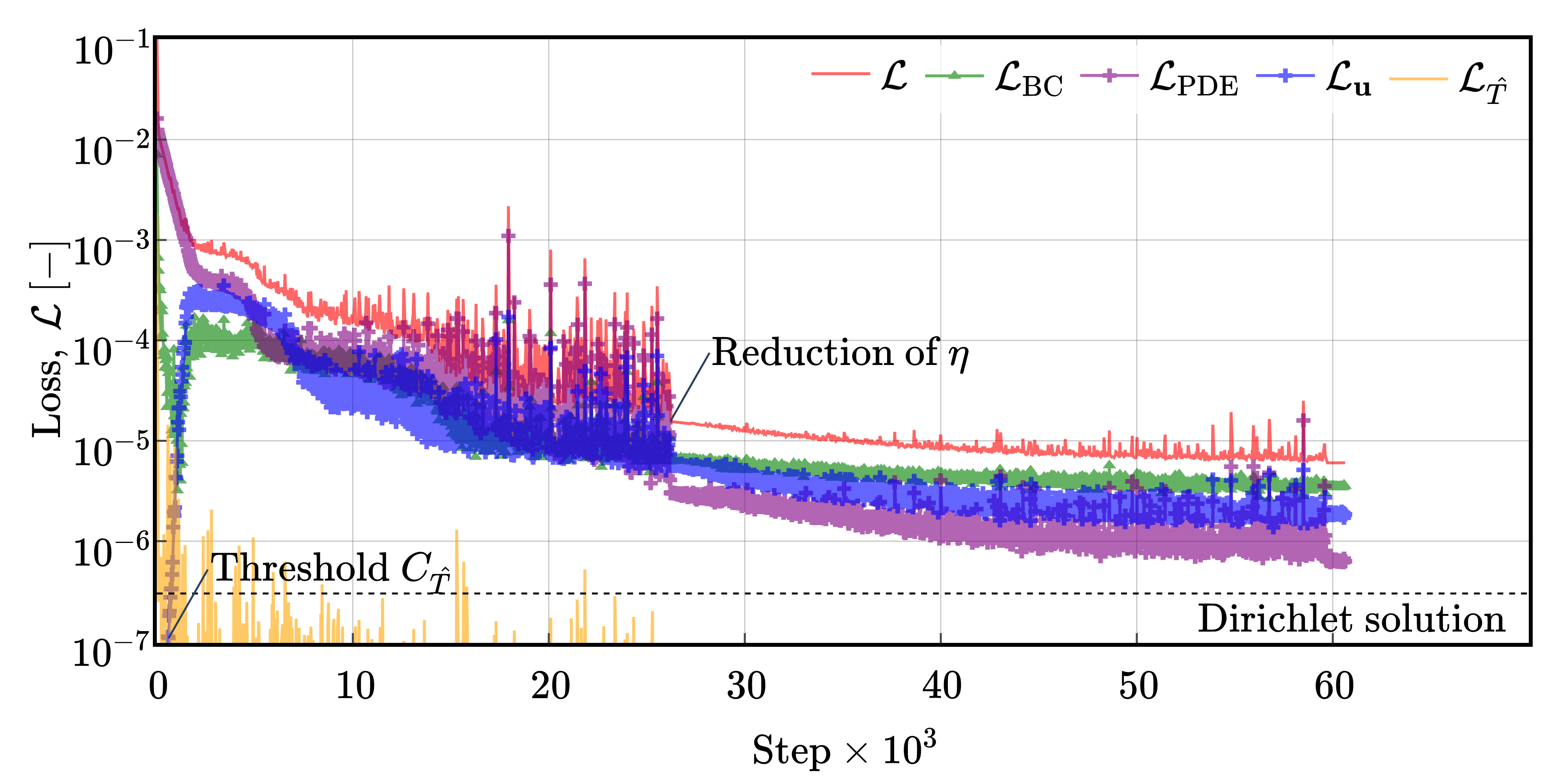}
    \caption{
    Loss convergence of the \acrshort{2D} model. The plot illustrates the convergence of the total loss and its contributing components: boundary condition, PDE, experimental displacement, and penalty on negative losses. The Dirichlet solution is presented for reference.
    The loss is elevated compared to the 1D model and convergence is not as straight, \cf\ \ref{fig:paper_1_losses_1d_model}. The impact of adding the experiential loss is barely visible. 
    }
    \label{fig:paper_1_losses_2d_model}
\end{figure}

%
The main limitation of the \acrshort{2D} approach is the substantial increase in computational cost, attributed to a 100-fold increase in collocation points. However, training converged over approximately the same number of epochs as the \acrshort{1D} model. We investigate the training performance of a complete 2D model and the addition of the experimental nugget diameter $\nuggetDia$. The latter is the key benefit of the \acrshort{2D} extension. The same hyperparameters were used for the \acrshort{2D} model, and $\lambda_{\nuggetDia} = 10$ was set. The loss convergence is depicted in \fig~\ref{fig:paper_1_losses_2d_model}, showing loss curves for $\lossBC$, $\lossPDE$, $\loss_\posTS$, and $\loss_{\temperatureEst}$. The time series loss is activated about 2,000 epochs earlier than in \fig~\ref{fig:paper_1_losses_1d_model} and converges with two learning rate reductions and stops after 1,000 epochs of no improvement. Notably, while the \acrshort{1D} model exhibited a prominent loss spike, the \acrshort{2D} model did not, indicating a different optimization dynamic. The loss preventing negative temperatures essentially vanishes after 25,000 epochs. The magnitude of $\loss$ is higher than with the \acrshort{1D} model, but the resulting temperature field does not show unexpected values.

\subsubsection{Predicted dynamic displacement}
\begin{figure}
    \centering
    \includegraphics[width=\imgWidth\linewidth]{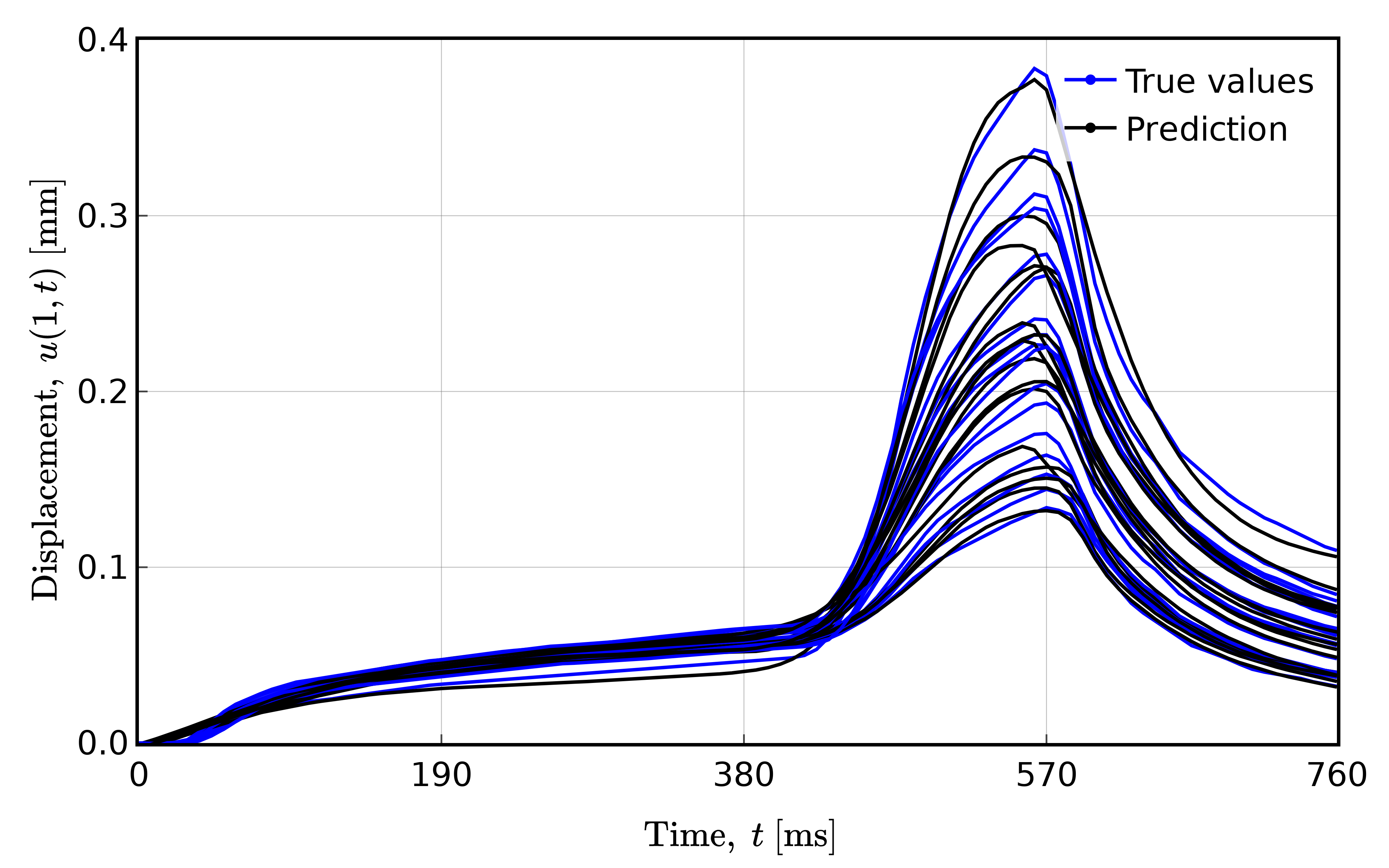}
    \caption{
    Comparison of experimental dynamic displacement to predictions of the 2D model. Discrepancies are observed during the initial phase, while closer agreement is found during the upslope and main welding phases.
    }
    \label{fig:paper_1_2d_model_all_test}
\end{figure}

Dynamic displacement curves are of particular interest in RSW for interpreting the nugget development, as they capture and measure the thermal response of the material to the welding process. Although much of the research on dynamic displacement focuses on steel \autocite{Gedeon1986a,Gedeon1986b,Gould1987,Lane1987,Wei1990,Dickinson1990}, the stages of relating to the nugget development identified for steel can be adapted for aluminum. However, aluminum's lower melting point, the absence of a zinc coating, or its higher thermal conductivity alter the timing and characteristics of the stages. A comparison of the 1D and 2D models reveals the trade-off of modeling aluminum RSW with a reduced or more complex model.
The 1D model predictions on the test data are visualized with the predicted temperature field in \fig~\ref{fig:visualize_full_model_low} and \fig~\ref{fig:visualize_full_model_high}. The temperature distribution is continuous despite contact heat being a scalar value on the faying surface. This shows that the PDE equations successfully enforce a continuous temperature field.
The results from the 2D model are shown in \fig~\ref{fig:paper_1_2d_model_all_test}, consisting of 17 test displacement curves from hold-out data. In this figure, the dotted line represents the prediction, while the solid line represents the experimental data.
The general trend is well fit, but two observations are visible. First, the initial part is not well fitted, between 0 and 100 ms. Second, the prediction is smoother and mismatches slightly, approaching the curve peak. Some curves exhibit unexplained worse predictions, indicating potential adverse experimental conditions or model accuracy.

Apart from occasional molten material, no nugget development is expected during stage I. This agrees with the 1D model, which predicts the first heating during preheating until $\zeit = 400$. The heat generation is visible for low forces, \eg\ \fig~\ref{fig:visualize_full_model_low}. Apart from this effect, $\posTS$ is unaffected during preheating because it is fixed at 12 kA during this time. 
Stage II is defined by a rapid increase in thermal expansion upon the point where the melting point is reached. This linear increase is visible in $\posTS$ for both models.
Stage III marks the slope decrease of $\posTS$ and the beginning of a saturation due to nugget development. This becomes visible once $\liqTemp$ is reached in the temperature field. This means that the end of the linear slope marks the beginning of the nugget development in aluminum RSW. The temperature field indicates the exact timing of nugget formation, as described in stages III and IV. The 1D model predicts the displacement within the confidence interval, but the model systematically overestimates this turning point. The 2D model exhibits a closer fit at this stage. The last part of the welding schedule, the hold time, cools the sheet's surfaces.
Our experiments do not exhibit excessive heat generation or spatter, characteristic of stage IV for steel RSW.

\subsubsection{Predicted nugget diameter development}
\begin{figure}
    \centering
    \includegraphics[width=\imgWidth\textwidth]{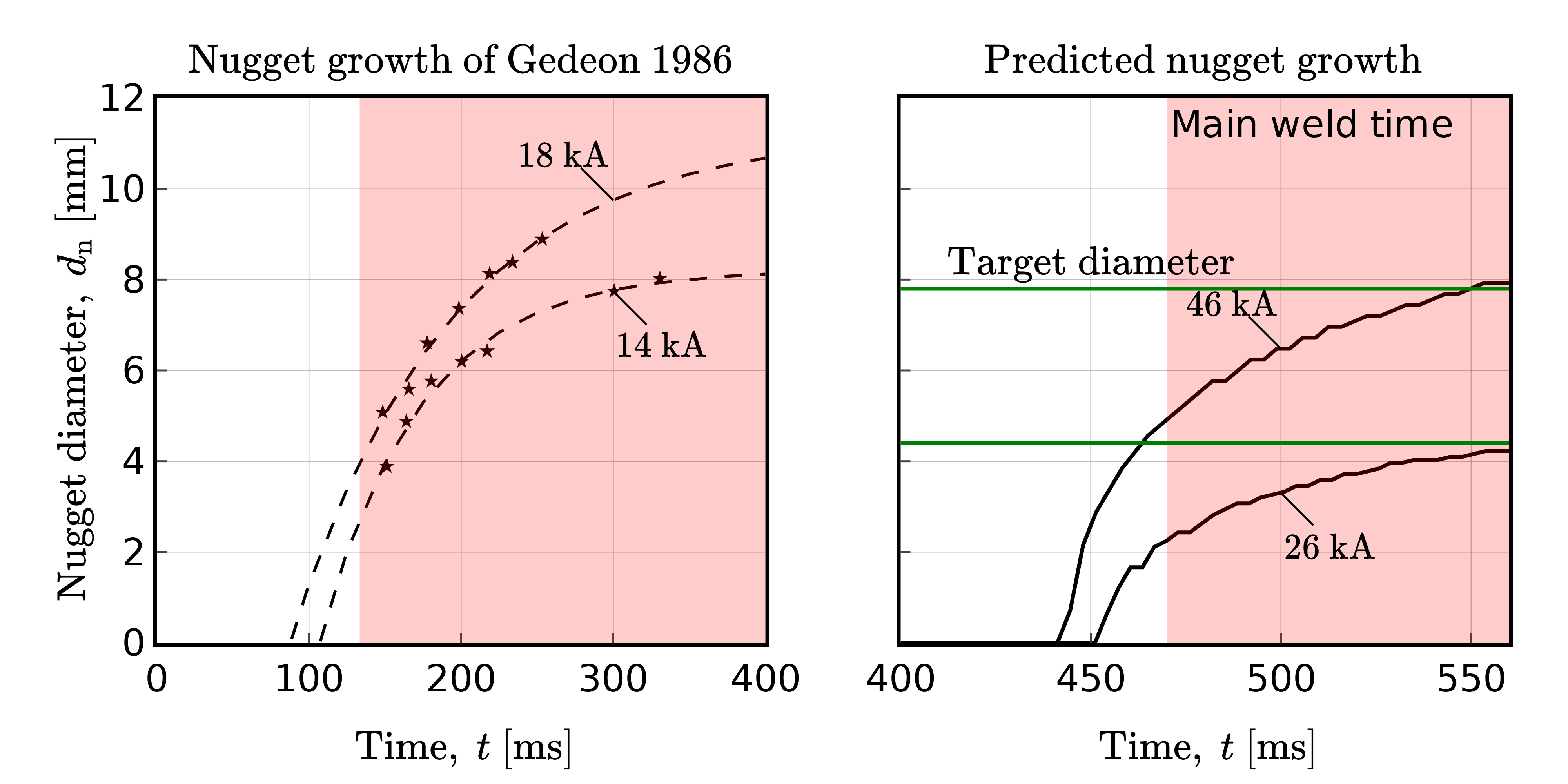}
    \caption{
    Comparison of the predicted nugget development and data for steel from \textcite{Gedeon1986b}. The qualitative shape of both is observed to be comparable.
    }
    \label{fig:paper_1_nugget_growth}
\end{figure}
\begin{figure}
    \centering
    \includegraphics[width=\imgWidth\linewidth]{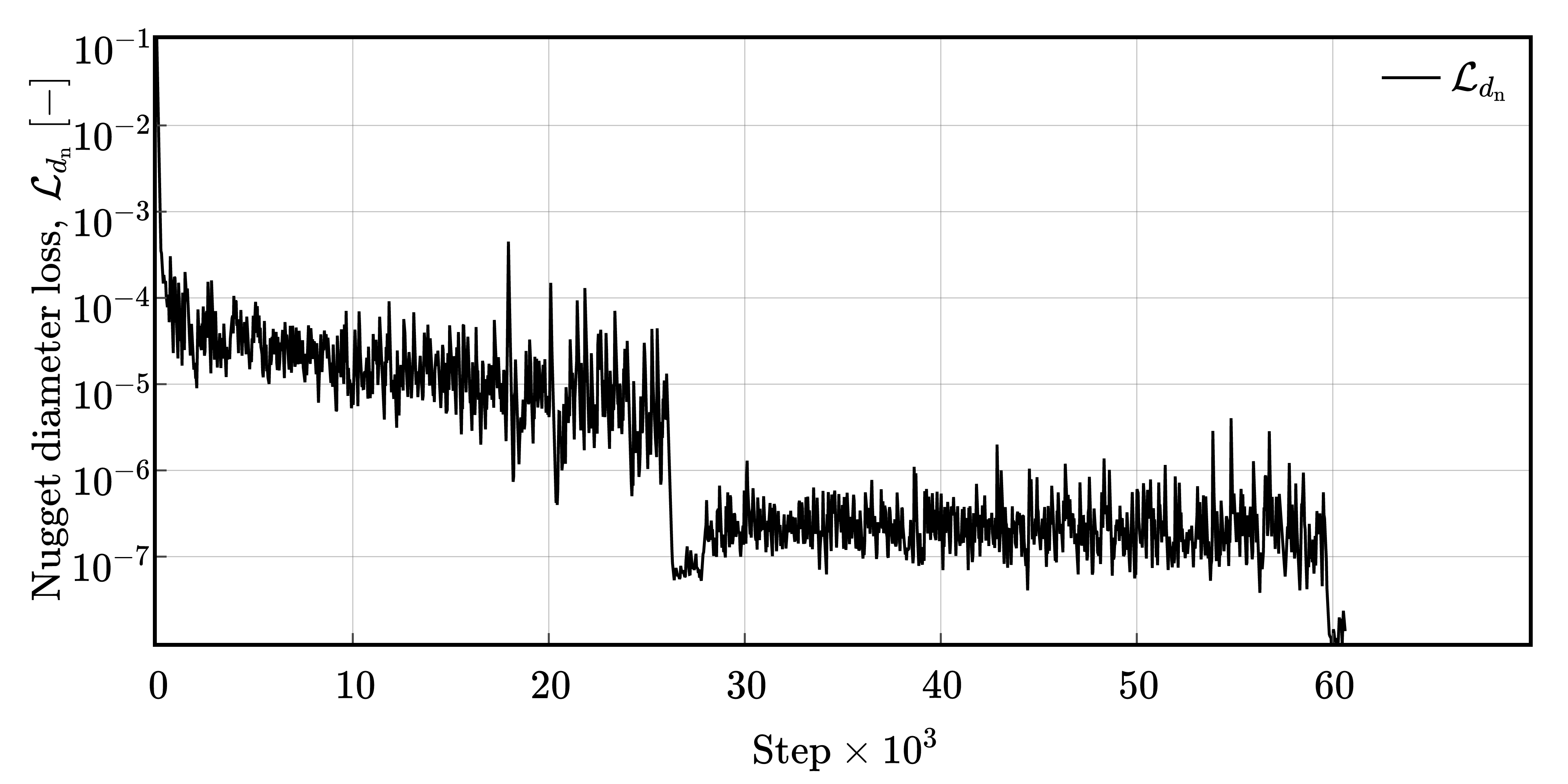}
    \caption{
    Loss convergence of the nugget diameter loss $\loss_{\nuggetDia}$ of the 2D model.
    }
    \label{fig:paper_1_nugget_loss}
\end{figure}

While FEM models \autocite{Hou2007,Tsai1991,Eisazadeh2010,Saleem2012} offer valuable insights, they differ from machine learning approaches in that they do not learn from data. 
In contrast, an advantage of PINNs is their ability to generalize between datapoints, effectively learning from experimental data while adhering to underlying physical relationships.
The \acrshort{2D} PINN can predict the complete temperature field $\temperatureEst(r, z, \zeit)$ on $\datasetTest$, including time steps towards the end of the main welding time, $\zeit = 560$, which are essential for the nugget diameter development.
The performance is quantified by the loss convergence of $\loss_{\nuggetDia}$ and fit on experimentally measured nugget diameters of $\datasetTest$. The nugget diameter loss penalizes predicted temperature below the liquidus at $\temperature(r,z,\zeit=560)$, \cf\ \equ~\ref{eq:goal_loss_dp}. This goal loss is tracked during training and depicted in \fig~\ref{fig:paper_1_nugget_loss}. It convergence to $\loss_{\nuggetDia} < 10^{-7}$, indicating no optimization problem for the PINN.

\begin{figure}
\centering
    \includegraphics[width=0.9\textwidth]{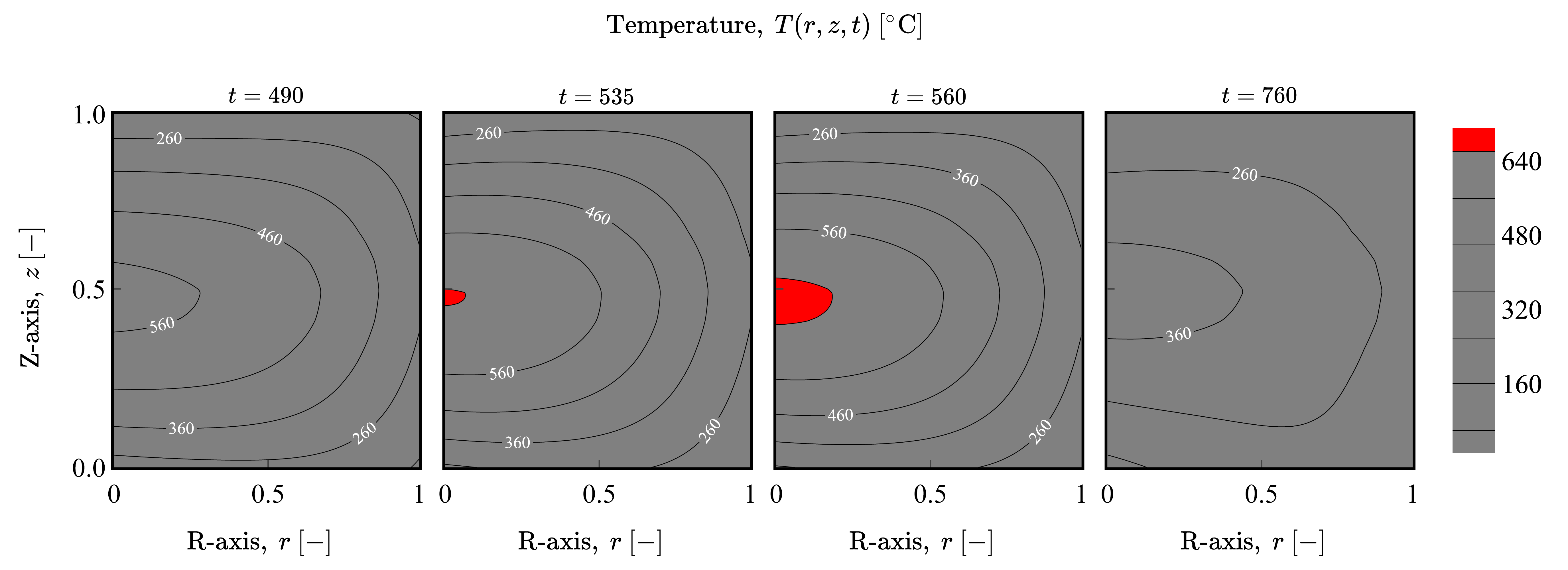}
    \caption{Two-dimensional temperature field $\temperature(r,z,t)$ with $\currentMax = 26$ and $\forceMax = 7$.
    }
    \label{fig:pinn_2d_pred_1}
\end{figure}

\begin{figure}
\centering
    \includegraphics[width=0.9\textwidth]{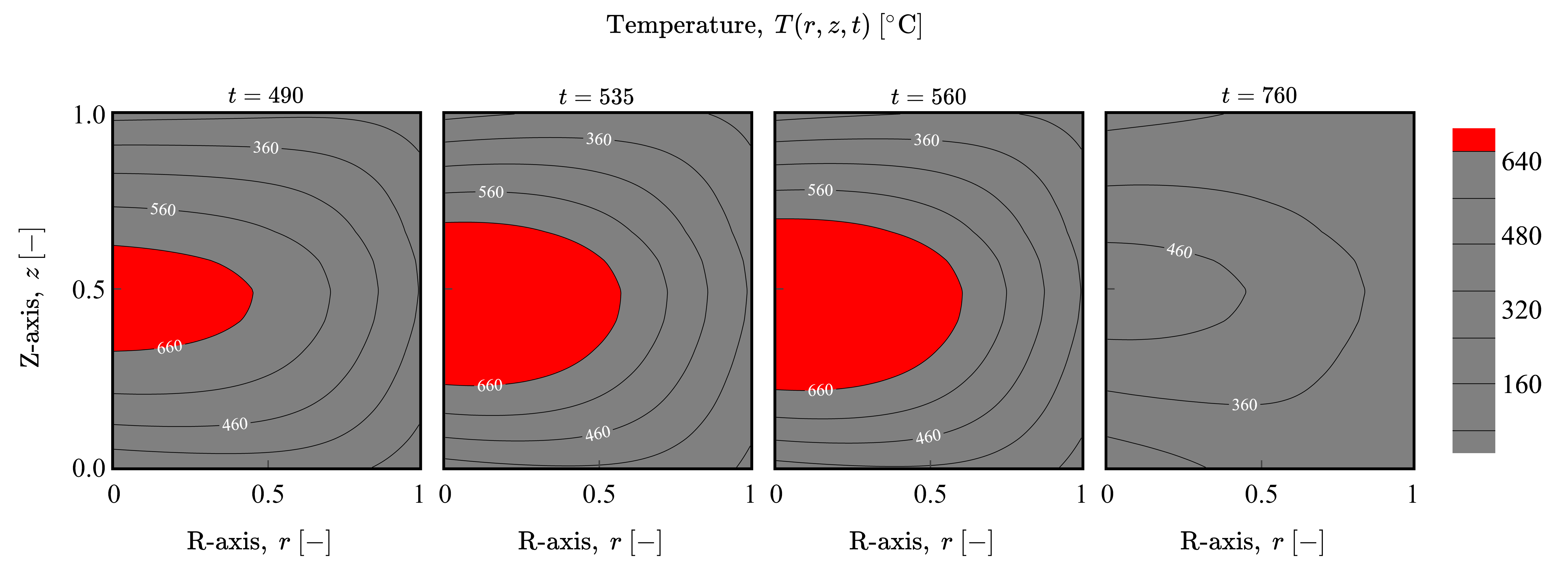}
    \caption{Two-dimensional temperature field $\temperature(r,z,t)$ with $\currentMax = 32$ and $\forceMax = 5$.
    }
    \label{fig:pinn_2d_pred_2}
\end{figure}

\begin{figure}
\centering
    \includegraphics[width=0.9\textwidth]{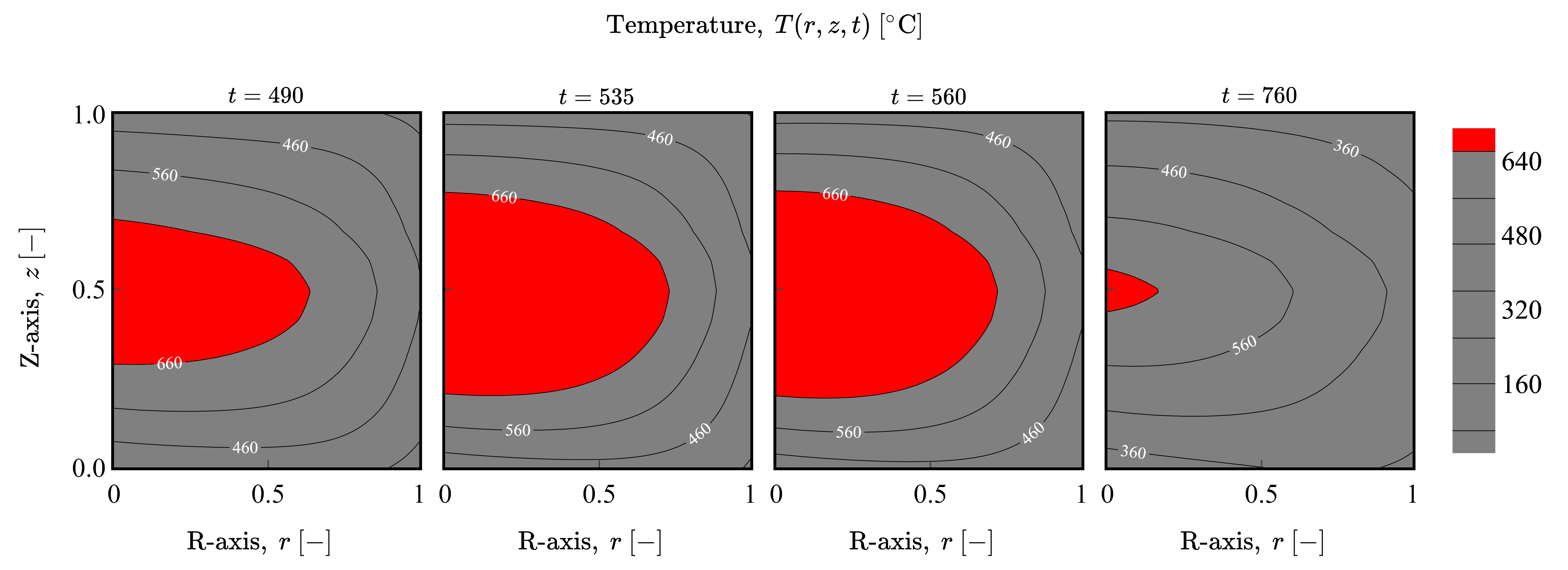}
    \caption{Two-dimensional temperature field $\temperature(r,z,t)$ with $\currentMax = 43$ and $\forceMax = 5$.
    }
    \label{fig:pinn_2d_pred_3}
\end{figure}

\textcite{Gedeon1986b} provided experimental data on the nugget growth of galvanized steel. The authors used a similar welding schedule with upslope and main weld time, albeit with lower durations and current levels. Despite the differences in material, the key trend in nugget development is assumed to be comparable. We compare this nugget development by predicting the 2D temperature field and counting mesh points exceeding the liquidus at the faying surface $\sheetSheetBC$ for every $\zeit$.
\fig~\ref{fig:paper_1_nugget_growth} shows the nugget development of galvanized steel on the left and aluminum on the right. We fit Gedeon's data with a curve for visual aid. Because of the mesh granularity, some discontinuities are present.
This comparison illustrates the material difference. For steel, a four-kiloampere difference over 266 milliseconds resulted in a 2 mm difference in nugget diameter. Meanwhile, aluminum RSW used a 20-kiloampere difference over 90 milliseconds to increase four millimeters. This difference stems from increased thermal conductivity, which is why higher amperages and shorter welding times are typically used. The PINN learns to predict the nugget diameter with acceptable accuracy. The nugget development is visually comparable to steel RSW. The liquidus temperature is reached during the upslope. In the following milliseconds, the nugget develops rapidly, followed by a period of stagnation during the main welding time. This rapid, linear growth corresponds to the dynamic displacement stage II, followed by a stagnation during stage III. With \acrlong{AL6} we do not observe a stage-IV-expulsion.
The nugget prediction is visualized for the full temperature field at selected $t$ for different $\currentMax$ and $\forceMax$ in Figs.~\ref{fig:pinn_2d_pred_1}, \ref{fig:pinn_2d_pred_2}, and \ref{fig:pinn_2d_pred_3}. We observe that the temperature field is slightly asymmetric. 
\fig~\ref{fig:pinn_2d_pred_3} estimates remaining liquid material, evident in the time step $\zeit = 760$. In general, the weld nugget forms uniformly and in the form of an ellipse. This was expected but not prescribed in the model.
\section{Conclusion}
%
We model aluminum \acrshort{RSW} using \acrshort{PINN}s, integrating experimental data and governing equations within a single model. The governing equations and boundary conditions were defined to train a \acrshort{PINN} to predict the displacement and temperature fields of aluminum RSW. Special attention is directed towards the welding control behavior, as it affects the displacement measurements and contact heat generation, which is important for RSW. Experimental data, including material parameters, dynamic displacement, and nugget diameter, were integrated using training strategies to achieve rapid loss convergence.
A \acrshort{1D} model is developed and used to investigate the training strategies. The model captures the dynamic displacement within the experimental variance for various process parameters. The experiments are faster in one dimension as the curse of dimensionality becomes apparent in the second dimension.
The \acrshort{2D} model includes the nugget diameter at the end of the main welding time as an optimization criterion. Therefore, it learns to represent the dynamic displacement and nugget diameter in terms of the displacement and temperature fields. The typical dynamic displacement stages of steel RSW are transferred to the PINN prediction.
Regarding the model training, relatively small networks were able to yield fast loss convergence and acceptable accuracy.
This efficiency, particularly in the one-dimensional model, makes it valuable for predictive control in industrial applications due to its instant predictive capability. Furthermore, an advantage of the model is its ability to generalize effectively across different datapoints, making it robust for varied experimental conditions.
In conclusion, our study demonstrates that integrating experimental data with governing equations through PINNs effectively predicts displacement and temperature fields in aluminum RSW. The resulting reference curves offer valuable insights into the process dynamics. This approach highlights the potential for optimizing welding parameters and the corresponding reference curve using \acrshort{PINN}s and guides future research.

\clearpage
%

%
%

%
%
\clearpage
\small

\printbibliography[title=References]

\end{document}
